\documentclass[conference]{IEEEtran}
\usepackage{times}

\usepackage[numbers]{natbib}
\usepackage{multicol}
\usepackage[pagebackref=true,breaklinks=true,bookmarks=true,colorlinks]{hyperref}
\usepackage[cal=cm]{mathalfa}
\usepackage{graphics} 
\usepackage{epsfig} 
\usepackage{mathptmx} 
\usepackage{times} 
\usepackage{amsmath} 
\usepackage{amssymb}  
\usepackage{siunitx} 
\usepackage{textcomp}
\usepackage{gensymb} 
\usepackage{booktabs}
\usepackage{multirow}
\usepackage{xcolor}
\usepackage{algpseudocode}
\usepackage{algorithm}

\let\labelindent\relax
\usepackage{enumitem}
\usepackage{xspace}
\setlist{nolistsep}

\usepackage{makecell}
\usepackage{multicol}
\usepackage{rotating}
\usepackage[percent]{overpic}
\usepackage{contour}
\usepackage{courier}

\usepackage{subcaption} 
\usepackage[font=small]{caption}
\usepackage[percent]{overpic}

\definecolor{MyDarkBlue}{rgb}{0,0.08,1}
\definecolor{airforceblue}{rgb}{0.36, 0.54, 0.66}
\definecolor{MyDarkGreen}{rgb}{0.02,0.6,0.02}
\definecolor{MyDarkRed}{rgb}{0.8,0.02,0.02}
\definecolor{MyDarkOrange}{rgb}{0.40,0.2,0.02}
\definecolor{MyPurple}{RGB}{111,0,255}
\definecolor{MyRed}{rgb}{1.0,0.0,0.0}
\definecolor{MyGold}{rgb}{0.75,0.6,0.12}
\definecolor{MyDarkgray}{rgb}{0.66, 0.66, 0.66}
\definecolor{MyPink}{rgb}{0.9, 0.33, 0.5}

\newcommand{\mypara}[1]{\par\vspace*{0mm} \textbf{{#1}}}

\newcommand\blfootnote[1]{%
  \begingroup
  \renewcommand\thefootnote{}\footnote{#1}%
  \addtocounter{footnote}{-1}%
  \endgroup
}

\def\OURS{RoboNinja}
\def\OURSNA{Non-Adaptive}
\def\HEURISTIC{Greedy}
\def\NN{NN}

\def\RL{RL}

\begin{document}

\title{RoboNinja: Learning an Adaptive Cutting Policy for Multi-Material Objects} 

\author{
Zhenjia Xu$^1$,\quad
Zhou Xian$^2$,\quad
Xingyu Lin$^3$,\quad
Cheng Chi$^1$,\quad
Zhiao Huang$^4$,\\
Chuang Gan$^{5\dagger}$,\quad
Shuran Song$^{1\dagger}$ \\
$^1$ Columbia University\quad
$^2$ CMU\quad
$^3$ UC Berkeley\quad
$^4$ UC San Diego \quad
$^5$ UMass Amherst \& MIT-IBM AI Lab
\\ \href{https://roboninja.cs.columbia.edu/}{https://roboninja.cs.columbia.edu/}}


%

\maketitle
\blfootnote{$\dagger$ indicates equal advising.}
\begin{abstract}
We introduce RoboNinja, a learning-based cutting system for multi-material objects (i.e., soft objects with rigid cores such as avocados or mangos). In contrast to prior works using open-loop cutting actions to cut through single-material objects (e.g., slicing a cucumber), RoboNinja aims to remove the soft part of an object while preserving the rigid core, thereby maximizing the yield. To achieve this, our system closes the perception-action loop by utilizing an interactive state estimator and an adaptive cutting policy.  The system first employs sparse collision information to iteratively estimate the position and geometry of an object's core and then generates closed-loop cutting actions based on the estimated state and a tolerance value. The ``adaptiveness'' of the policy is achieved through the tolerance value, which modulates the policy's conservativeness when encountering collisions, maintaining an adaptive safety distance from the estimated core. Learning such cutting skills directly on a real-world robot is challenging. Yet, existing simulators are limited in simulating multi-material objects or computing the energy consumption during the cutting process. To address this issue, we develop a differentiable cutting simulator that supports multi-material coupling and allows for the generation of optimized trajectories as demonstrations for policy learning. Furthermore, by using a low-cost force sensor to capture collision feedback, we were able to successfully deploy the learned model in real-world scenarios, including objects with diverse core geometries and soft materials. 
\end{abstract}


\IEEEpeerreviewmaketitle

\section{Introduction}

\label{sec:introduction}
\begin{figure}[t]
    \centering
    \includegraphics[width=\linewidth]{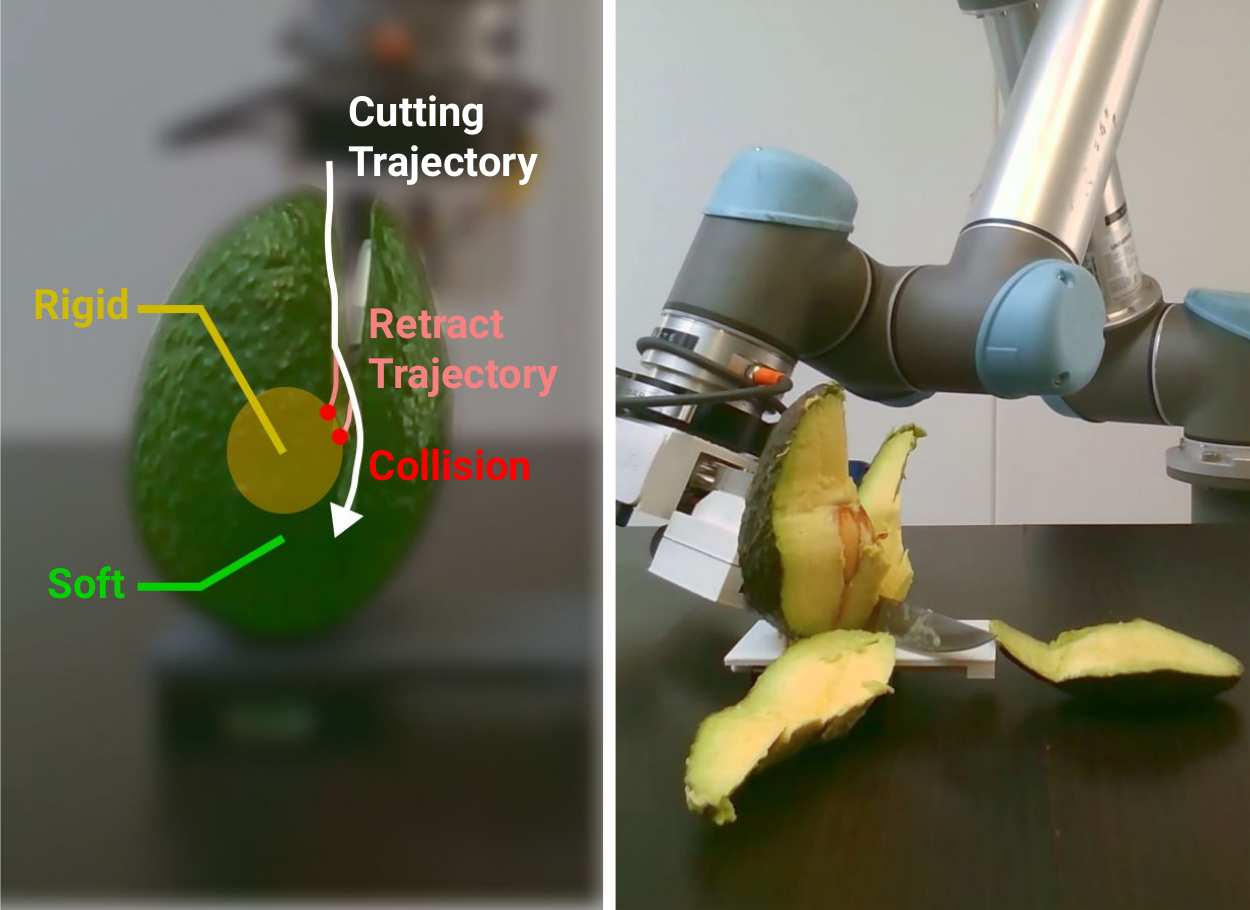}
    \caption{\textbf{\OURS} is designed to cut multi-material objects with an interactive state estimator and adaptive cutting policy. Left: When the knife encounters a collision with the invisible core, the algorithm updates the core estimation and re-plans the cutting trajectory after a few retracting actions. Right: We deploy the learned  model on a physical robot, allowing it to cut fruits in a way that maximizes the cut-off mass while minimizing collision occurrences.}
    \label{fig:teaser}
    \vspace{-5mm}
\end{figure}

Imagine slicing a piece of avocado from its seed (Fig. \ref{fig:teaser}) -- we need to carefully slice through the soft outer flesh to locate the rigid seed and then follow the contours of the seed to maximize the volume of the slice. In some cases, we would need to switch the cutting trajectory when the knife collides with the seed. 
All of these maneuvers must be performed while adhering to the physical constraints of the knife and its interactions with both the soft and rigid parts of the avocado.  

This simple yet intricate task illustrates the challenges of cutting multi-material objects, which is significantly more difficult than cutting through single-material objects that often could be accomplished with an open-loop cutting trajectory \cite{heiden2021disect, long2013robotic, sharma2019learning, mu2019robotic,zhou2006cutting1, zhou2006cutting2}. 
In this paper, we are interested in enabling robots to effectively and efficiently perform this task.
Using the above example, we could summarize the unique capabilities required for acquiring such a skill: 
\begin{itemize}[leftmargin=5mm]
    \item \textbf{Multi-objective optimization under complex physical constraints.} To perform the task, the system needs to simultaneously optimize several objectives -- maximizing the total yield (cut-off mass of the soft material), avoiding collisions with the rigid core, and minimizing energy consumption. Many of these objectives require comprehensive physical reasoning beyond simple geometry analysis.  

    \item \textbf{Interactive state estimation for extreme partial observability.} In most cases, the rigid core is not observable on the surface. Hence, it requires the system to continuously estimate the location and geometry of the core through interaction, that is, through cutting and sensing the collision. This state estimator will continuously inform the cutting policy in a close-loop manner.

    \item \textbf{Adaptive policy for out-of-distribution scenarios.}  While the state estimator could infer the core geometry based on contacts, there will always be instances where the shape of the core falls outside the training distribution. An inaccurate estimation could lead to repetitive collisions in the same location. In these scenarios, the cutting policy needs to ``adaptively'' update its cutting strategies to avoid getting stuck.
\end{itemize}

As the first step towards enabling this new robot capability, we introduce \textbf{\OURS}, a learning-based cutting system that combines an interactive state estimator and an adaptive cutting policy. 
The interactive state estimator uses sparse contact information to iteratively estimate the position and shape of the core. 
The cutting policy, optimized to increase cut-off yield and reduce collision occurrences and energy consumption, produces cutting actions in a closed-loop manner, based on the estimated state and a tolerance value. The tolerance value is a function of past collision events and actively controls the policy conservativeness when encountering new collisions (e.g., keeping a distance from the estimated core location). This adaptivity is critical for handling out-of-distribution scenarios where the state estimation could be inaccurate.

Learning such cutting skills directly on a real-world robot system is challenging and potentially dangerous. However, existing simulators in the literature are limited in simulating multi-material objects, especially the coupling between rigid and soft bodies under forceful manipulations such as cutting. Therefore, we develop a new differentiable simulator for the proposed multi-material object cutting problem, allowing us to use gradient-based optimization for generating trajectories as demonstrations for policy learning. 

Finally, when deploying the learned policy, we demonstrate that with the simple collision feedback captured by a low-cost (less than \$10) force sensor, we can successfully transfer the model learned in the simulator directly to real-world scenarios, including out-of-distribution object geometries and materials, thanks to its adaptivity.

In summary, the primary contribution of this paper is \OURS~-- the first robotic system demonstrating the capability of multi-material object cutting. To build this system, we make the following technical advancements: 
\begin{itemize}[leftmargin=5mm]
\item The formulation of the multi-material cutting task and a differentiable simulator that could perform multi-objective trajectory optimization for collecting demonstrations.

\item A learning-based cutting method with an \textit{interactive} state estimator and an \textit{adaptive} cutting policy.

\item The deployment of our method on a real-world robotic system with low-cost sensory feedback. 
\end{itemize}

Videos of experiments, the code for the simulator and the cutting system, as well as the CAD models for the benchmark objects are available at \url{https://roboninja.cs.columbia.edu/}.
\begin{figure*}[t]
    \centering
    \includegraphics[width=\linewidth]{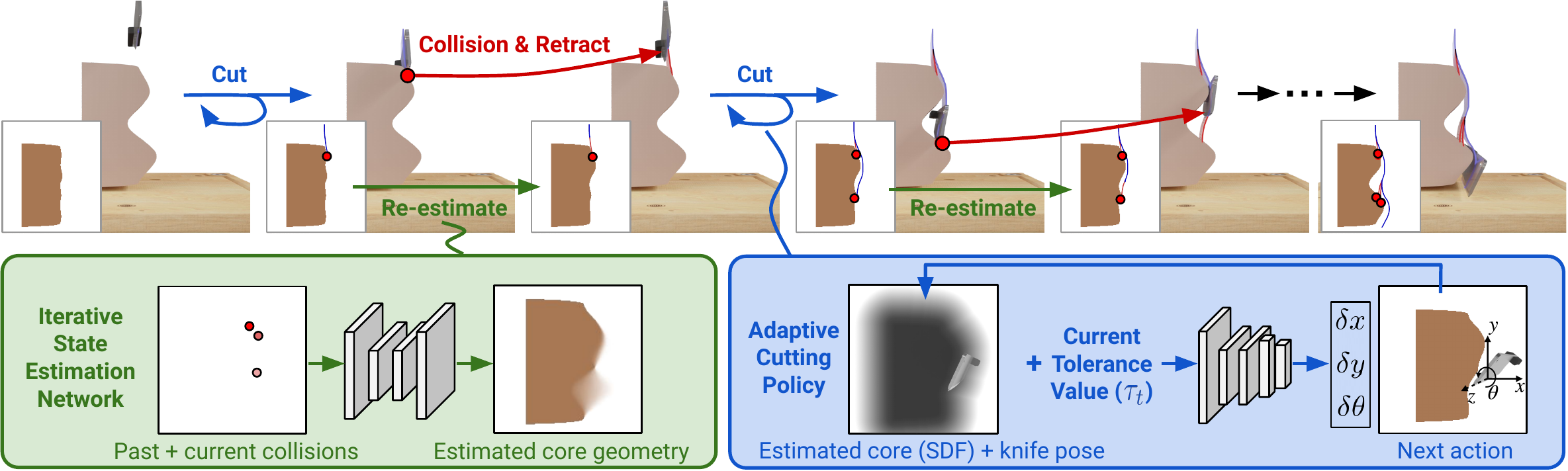}
    \caption{\textbf{\OURS~Overview.} The robot initially performs actions based on an initial estimation. In the event of a collision with the rigid core, the robot retracts a few steps (indicated in red). The robot then re-estimates the core's position and geometry (indicated in green) and generates the cutting action using the updated state estimation (indicated in blue).}
    \vspace{-2mm}
    \label{fig:method}
\end{figure*}

\section{Related Work}
\label{sec:related_work}

\subsection{Differentiable Physics Simulation for Policy Learning}
In recent years, a number of differentiable simulation environments have been proposed to accelerate policy learning. These include 1) simulators parameterized by neural networks \cite{li2019propagation, li2018learning, pfaff2020learning}, which haven't proved to be capable of accurate simulations involving complex interactions between multi-phase materials required in cutting scenarios, and 2) analytical simulators implemented in a differentiable way, leveraging either automatic differentiation tools \cite{hu2019difftaichi} or analytical gradient computation rules \cite{hu2019chainqueen}. The latter is used to provide gradient information to accelerate policy search in various robotic tasks, including locomotion \cite{xu2022accelerated}, soft robot design \cite{wang2023softzoo}, soft body manipulation \cite{huang2021plasticinelab, xian2023fluidlab, qi2022dough}, etc. To handle realistic sensory input outside of the simulation environment, prior methods distill the knowledge learned in simulators into visuomotor policies that take images~\cite{lin2021diffskill} or point clouds~\cite{lin2022planning} as input. In contrast, the state of multi-material objects in our task, such as the shape of the rigid core inside, is not directly observable. As such, our method learns an interactive state estimation network based on encountered collision events to address the challenge.


\subsection{Simulation Environments for Cutting}
There has been a significant amount of research conducted on simulating the cutting process of different materials. Theoretical analysis is commonly applied in metal cutting \cite{merchant1945mechanics, childs2006friction} and brittle materials research \cite{griffith1921vi, miller1999modeling, zhou2005rate}. Besides, various numerical methods are utilized to simulate the fracture of deformable objects. They could be further classified into mesh-based methods, such as the Finite Element Method (FEM) \cite {heiden2021disect, areias2017steiner, koschier2014adaptive, wu2015survey, paulus2015virtual, jevrabkova2009stable}, and mesh-free methods, including Position-based Dynamics (PBD) \cite{pan2015real, berndt2017efficient} and Material Point Method (MPM) \cite{hu2018moving, wang2019simulation, wolper2019cd, wolper2020anisompm}.
The work most related to ours is ``DiSECt'' by Heiden et al. \cite{heiden2021disect}, where the FEM-based cutting simulator achieves differentiability through continuous contact formulation and damage modeling.
In contrast to these prior works that only simulate the cutting process of a single material, our differentiable simulator stands out by accounting for both the external soft material and the internal rigid core, utilizing MPM for its ability to accurately and efficiently simulate the dynamics of elastoplastic objects and the coupling between different materials. 

\subsection{Interactive Perception}
Interactive perception (IP) \cite{bohg2017interactive} leverages physical interaction to gather information about the environment. It is commonly used for scene reconstruction with occluded objects \cite{xu2020learning, kenney2009interactive, schiebener2013integrating, schiebener2011segmentation} and kinematic object structure discovery \cite{jiang2022ditto, gadre2021act, nie2022sfa, lv2022sagci}. The integration of tactile signals becomes increasingly popular in this field, especially in  material classification \cite{culbertson2014modeling, chu2015robotic}, object recognition \cite{liu2017recent, xu2022tandem, xu2022tandem3d, schmitz2014tactile, schneider2009object}, and shape reconstruction \cite{allen1990acquisition, bierbaum2008robust, matsubara2017active, lu2022curiosity}. Recently, Xu et al. \cite{xu2022tandem3d} recognize 3D objects with active tactile explorations. Wang et al. \cite{lu2022curiosity} present a curiosity-driven object reconstruction method using the modality of touch. These prior works only consider rigid or articulated objects. Our work advances the field of interactive perception by applying it to a new and challenging task: multi-material object cutting.

\subsection{Robotic Cutting}
Many robotic systems have been developed for cutting tasks in various domains such as meat \cite{long2013robotic}, vegetables \cite{sharma2019learning, mu2019robotic, zhou2006cutting1, zhou2006cutting2, sawhney2021playing}, and dough \cite{lin2021diffskill, lin2022planning}.
Researchers also utilize multimodal haptic sensory data to enhance system robustness \cite{zhang2019leveraging}. As an alternative to traditional cutting using knives, hot wire cutting tools \cite{duenser2020robocut, sondergaard2016robotic} are adopted in various sculpting and industrial applications.
Previous works mostly deal with only single-material objects and study how to cut through them, where open-loop orthogonal cutting is typically sufficient. In this work, we study the problem of cutting multi-material objects with the goal of removing soft material from an invisible rigid core, which requires controlling under physical constraints such as avoiding collisions with the rigid core,  and continuous state estimation from sensory feedback in a partially observable environment. Therefore, we propose to equip our system with an interactive state estimation network and an adaptive policy to address the task.


\section{Method}
\label{sec:method}

In this work, we study a multi-material object cutting task, where the goal is to manipulate a tool to cut off soft material from a rigid core while maximizing the yield (the total amount of soft material being removed), as well as minimizing the number of collisions with the core and the total energy consumption.

Considering the complex objectives and the real-time nature of this task, we decide to employ the standard teacher-student framework: slow while accurate expert demonstrations act as the teacher (optimized with the differentiable simulator), and a lightweight learning-based policy as a student is trained to imitate the teacher's behavior and is deployed at inference time. Specifically, we first build a physics-based differentiable simulator supporting multi-material coupling to gather expert demonstrations via gradient-based trajectory optimization. Next, we train an interactive state estimation network to infer the position and the geometry of the core based on collected collision signals.
Afterward, a cutting policy is trained to generate actions based on the core estimation and the knife state. This policy not only imitates the behavior of the expert demonstrations but also adaptively adjusts the conservativeness to retract from the core after the collision.
An overview of the execution is illustrated in Fig. \ref{fig:method}. The robot first starts the cutting process by executing actions based on an initial estimation from a learned prior. Upon collisions with the core, the state estimation is immediately updated. The robot subsequently retracts a few steps and replans the cutting trajectory using the updated state estimation. In this process, the system progressively updates the estimated state, leading to an accurate estimate of the position and geometry of the invisible core after multiple collision events, which in turn allows a physically plausible cutting trajectory to cut off most of the soft material.

In the following sections, we first present our expert demonstration process in \S \ref{method:trajectory_optimization}, where we develop our differentiable cutting simulator and perform gradient-based trajectory optimization. We then detail the iterative state estimation in \S \ref{method:state_estimation} and the adaptive cutting policy in \S \ref{method:cutting_policy}. Finally, we present a hardware setup that includes a low-cost force sensor for deploying our policy in real-world scenarios (\S \ref{method:real_setup}).

\subsection{Multi-objective Trajectory Optimization with a Differentiable Simulation Environment}
\label{method:trajectory_optimization}
\mypara{Differentiable cutting simulation environment} We build a cutting simulation environment to support the modeling of both soft and rigid materials, as well as the coupling between them. The soft material in our scene, e.g. flesh of fruit is represented using an elastoplastic continuum model simulated with MLS-MPM \cite{hu2018moving}, treated by the von Mises yield criterion. In contrast to FEM \cite{heiden2021disect}, MPM-based methods naturally support arbitrary deformation and topology change. For rigid bodies in the scene, including the rigid core, the knife, and the support surface (a chopping board), we represent them as time-varying signed distance fields (SDFs), converted from imported external meshes. We model the contact between soft and rigid materials by computing surface normals of the SDFs and applying Coulomb friction \citep{stomakhin2013material}. Material separation occurring during the cutting process is handled by MLS-MPM, which inherently supports modeling sharp and clean split of material points in the soft material. Our simulator is implemented in Python and Taichi. For gradient computation, we use both Taichi's autodiff system for simple operations and analytical gradient computation for complex operations such as SVD in deformation gradient updates. Additionally, in order to lift the computation bottleneck imposed by limited GPU memory size, we implement efficient gradient-checkpointing to allow gradient flow over long temporal horizons.

\mypara{Multi-objective trajectory optimization} At each step, we consider a cutting action parameterized by the knife orientation $\theta$ along the $z$ axis and a vertical displacement $\delta \vec{p} = (\delta x, \delta y)$ within the $x-y$ plane (see Fig. \ref{fig:method}). To further ensure the smoothness of the cutting trajectory, we impose an additional constraint on the magnitude of the displacement: $\|\delta \vec{p}\|=2mm$. Since our differentiable simulator allows gradient-based trajectory optimization, we collect cutting trajectories in the simulator assuming access to all ground truth information, optimized using the following objectives: 

\begin{figure}[t]
    \centering
    \includegraphics[width=\linewidth]{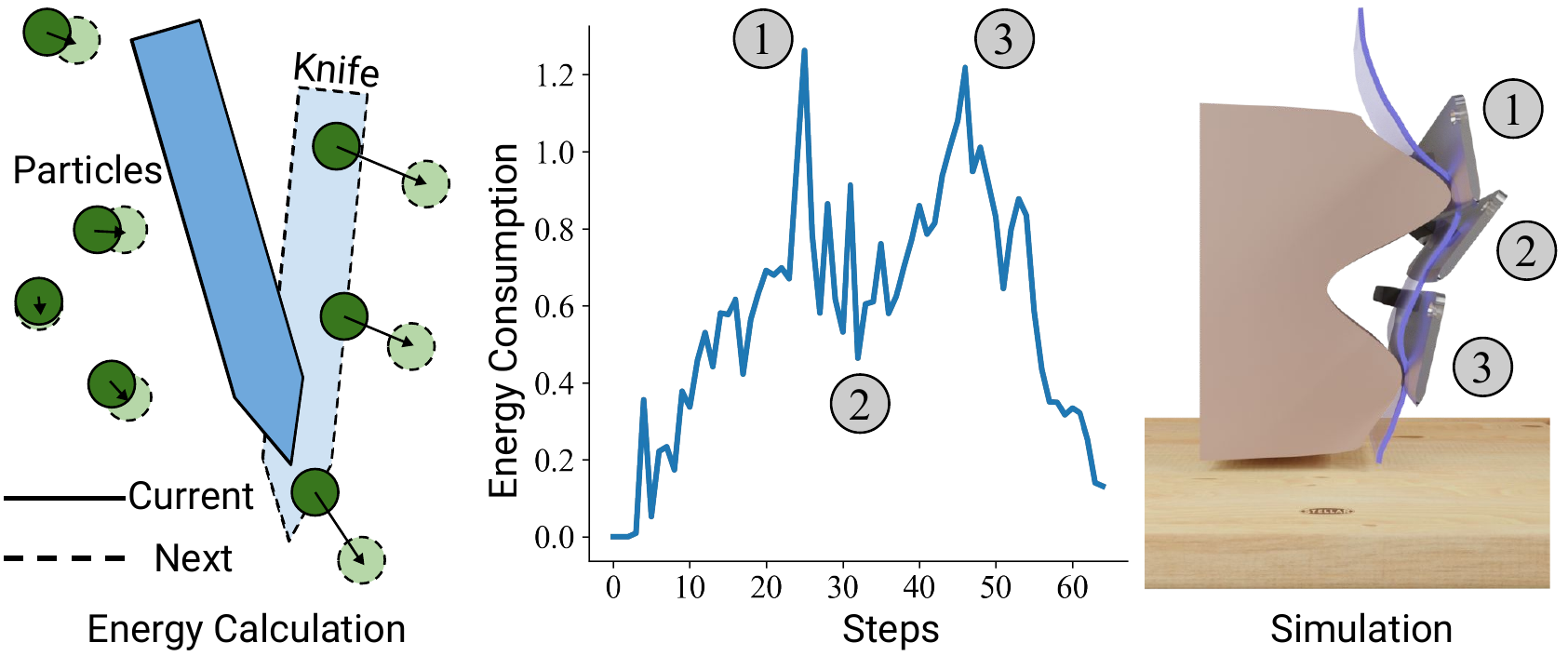}
    \caption{\textbf{Energy Calculation.} [Left] Illustration of energy computation in simulation. Particle position and velocity are changed due to collision with the knife. The cumulative work from the knife to each collided particle is considered as the energy consumption of the agent. [Middle] A plot of energy consumption at each step. [Right] Visualization of knife poses at some representative steps. Note that large energy consumption incurs around the rapid rotations in the trajectory. (e.g., steps 1 and 3).}
    \label{fig:energy_plot}
\end{figure}

\begin{itemize}[leftmargin=5mm]
    \item \textit{Cut mass}: one primary goal of a cutting policy is to maximize the total amount of the soft material being removed by a cutting trajectory. This objective $\mathcal{L}_m$ is computed by accumulating the mass of all material points removed from the rigid core at the end of each episode.
    \item \textit{Collision occurrences}: an ideal cutting trajectory should be able to avoid unnecessary collision events during its course. Our simulator represents both the core and the knife by time-varying SDFs. The collision between them is detected by sampling $N$ uniform points on the knife surface and checking whether they penetrate the core. In order to make this optimization process differentiable, we model the discontinuous contact between the knife and the core in a soft manner, following \cite{huang2021plasticinelab, xian2023fluidlab}, and compute a differentiable collision loss for optimization: $\mathcal{L}_{col} = \sum_{i=1}^N \|\text{max}(d_i+\hat{d}, 0)\|^k$, where $d_i$ represents the penetration distance of the $i$-th sampled point, and $\hat{d}$ is an additional safety margin. In practice, we found $N=5$, $k=4$, and $\hat{d}=2cm$ to be sufficient to produce good trajectories.
    
    \item \textit{Energy consumption}: in order to produce a natural and smooth motion trajectory during a cutting process, our system also optimizes energy consumption. We estimate the energy loss $\mathcal{L}_e$ based on the work done by the knife during its motion at each step, which is computed by summing the product of the distance traveled and force experienced by each material point in contact with the knife: $\mathcal{L}_e = \Sigma_j \frac{m_j\Delta \vec{v}_j}{\Delta t} \cdot \Delta{\vec{p}_j}$, where $\Delta \vec{v}_j$ and $\Delta \vec{p}_j$ denote the change in velocity and position for each particle $j$, respectively. In Fig. \ref{fig:energy_plot}, we illustrate the energy consumption of each step in an example cutting episode.
\end{itemize}
\begin{figure}[t]
    \centering
    \includegraphics[width=\linewidth]{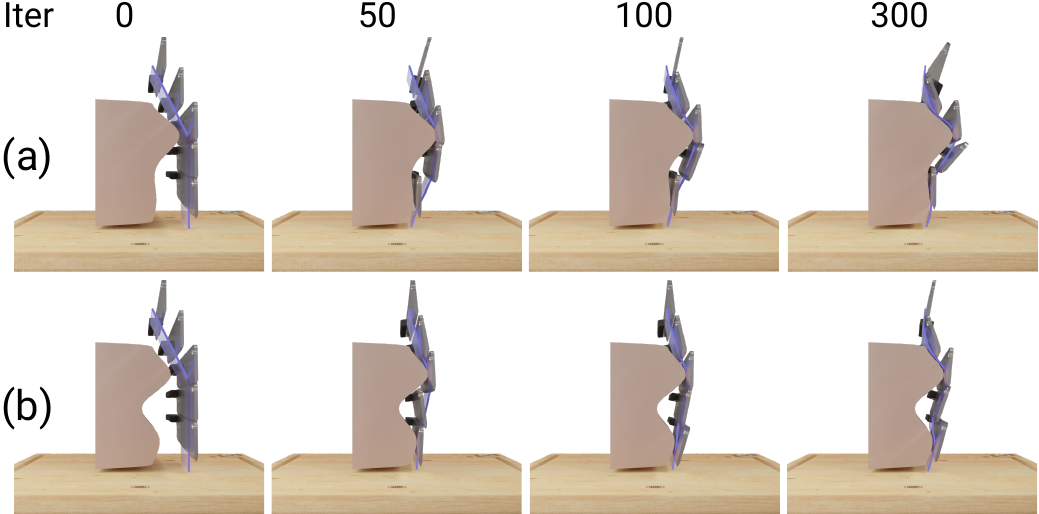}
    \caption{ \textbf{Optimization Process of Cutting Trajectory.} As illustrated in the first column, the cutting trajectory is initialized with a predesigned collision-free path. After 300 optimization iterations, the knife is able to cut off most of the soft materials with optimized energy consumption.}
    \vspace{-2mm}
    \label{fig:optimization_process}
\end{figure}
During trajectory optimization, we initialize using a pre-designed and translational collision-free trajectory that traverses down from above the object till touching the support surface, and optimize the trajectory loss by summing the above objectives: $\mathcal{L}_{total} = \mathcal{L}_m + \eta_{col}\mathcal{L}_{col} + \eta_{e}\mathcal{L}_{e}$, 
where $\eta_{col}$ and $\eta_{e}$ are coefficients for balancing different objectives. Examples of the optimization process are shown in Fig. \ref{fig:optimization_process}.



\subsection{Interactive State Estimation}
\label{method:state_estimation}
The purpose of the state estimation module is to determine the location and shape of the initially invisible core using collision signals collected during the cutting process.
As illustrated in Fig. \ref{fig:method}, the input is a sparse collision map within the $x-y$ plane, where each filled pixel represents a collision point encountered during the trajectory traveled until the current time step.
We employ an 11-layer U-Net architecture to estimate the 2D mask of the rigid core.

To facilitate offline training, we randomly select $k$ points on the contour of the training cores to mimic collision signals that might be received during actual execution. Here $k$ is an integer uniformly sampled from a uniform distribution with a range [0, 9]. The model is trained using the ground truth geometry as supervision and optimized with Binary Cross-Entropy loss. During the inference phase, a pre-determined threshold $S_{thr}$ is used to convert the predicted probability map to a binary core mask.

\subsection{Adaptive Cutting Policy}
\label{method:cutting_policy}




The goal of the cutting policy is to generate the cutting action at each step. 
The policy network takes the signed distance field of the estimated core mask, the current knife pose, and an additional tolerance value $\tau_t$ as input. It predicts the cutting action $a_t$, parameterized by the translational $\delta \vec{p} = (\delta x, \delta y)$ and rotational movement $\delta\theta$ of the knife.
Tolerance value $\tau_t$ controls the level of the conservativeness of the action. A lower tolerance value results in the knife getting closer to the estimated core, thus increasing the risk of collision. Conversely, a higher tolerance value leads to a more conservative action, keeping the knife farther from the core to prevent potential collisions. We initialize the tolerance at each step as 0.
Upon collisions, the robot first retracts the knife for $R_{dis}$ steps based on the history of actions, then executes actions with an increased tolerance value by $\tau^{+}$ for $R_{dis}$ step, and then linearly decay the tolerance $\tau$ afterwards to produce a smooth trajectory.


We use the demonstrations $\{D\}$ collected in the differentiable simulator to train our cutting policy. The tolerance value of the actions in the demonstration is assumed to be 0, and we use data augmentation to generate training data for tolerance values $\tau$ larger than 0 in the following way: For each demonstration, $D_i=[s_0, a_0, s_1, a_1, \cdots, s_N, a_N, s_{N+1}]$, where $s_i$ and $a_i$ represents the knife pose and action at each step, we generate another knife trajectory $\hat S = [\hat s_0, \hat s_1, \cdots, \hat s_N, \hat s_{N+1}]$ by moving the current knife trajectory away from the core in the direction of the x-axis by $\tau$.
To increase training robustness, a gaussian noise is added to the action sequence.
Finally, the quadruple $(\tau, s^*_0, a_i, \hat s_{i+1})$, along with the core geometry is used for training, where $s^*_0$ is calculated analytically using $a_i$ and $\hat s_{i+1}$. The cutting policy is trained with Mean Squared Error (MSE) loss.

\begin{figure}[t]
    \centering
    \includegraphics[width=\linewidth]{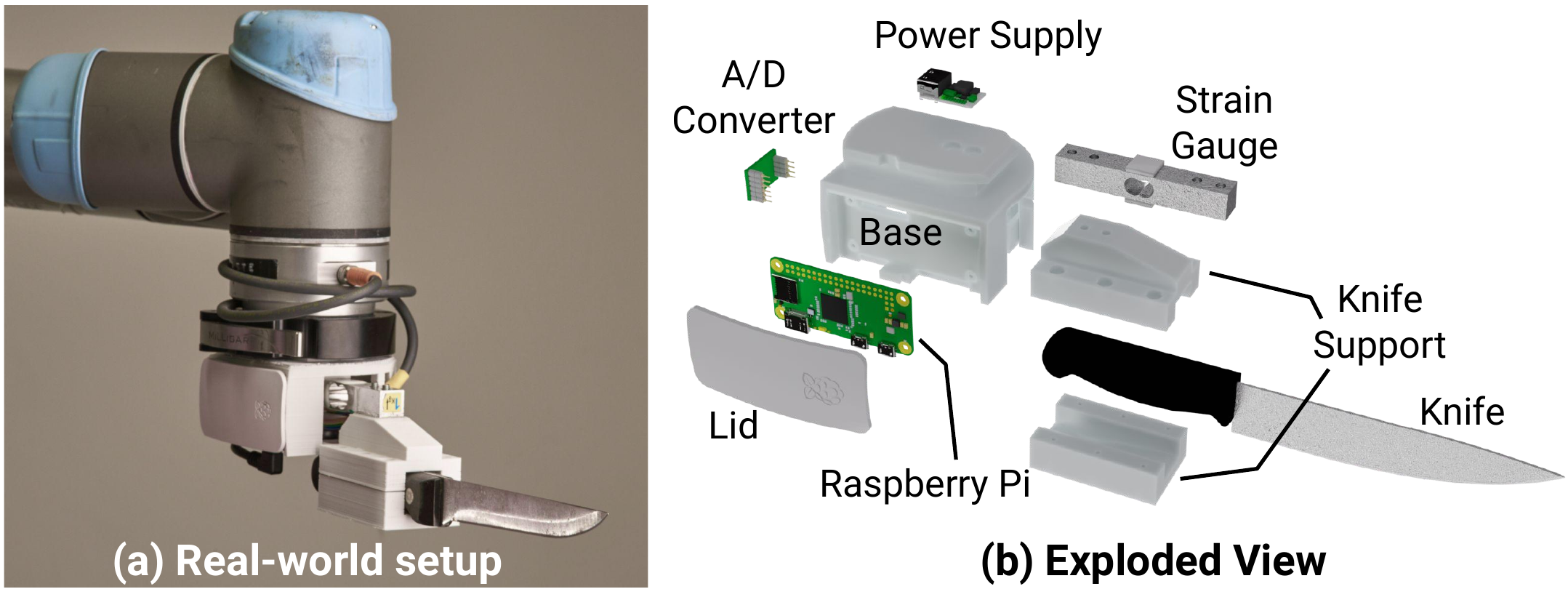}
    \caption{ \textbf{Hardware.} We design and construct a compact cutting tool equipped with a force sensor. The force is measured by a strain gauge as an analog electrical signal. The signal is then converted to a digital signal and transmitted to the robot controller through an A/D converter and a Raspberry Pi Zero, respectively.}
    \vspace{-4mm}
    \label{fig:hardware}
\end{figure}
\subsection{Real-world System Setup}
\label{method:real_setup}
We design and build a low-cost, force feedback system for deploying our method on a real-world UR5 platform. Figure \ref{fig:hardware} shows an image and the exploded view of our hardware system. 
A strain gauge load cell measures the shear force experienced by the connected knife as an analog signal. This signal is amplified and converted into digital readings by an HX711 A/D converter at 80 Hz. The digital measurement of the shear force is then processed by the Raspberry Pi Zero and transmitted to the robot controller via Wi-Fi. The entire system is powered by UR5's tool I/O power, and all these components are compactly assembled inside a 3D-printed container. The load cell and the AD converter are the core components of our hardware system to achieve real-time force feedback and are priced at   \$8 in total\footnote{Amazon link: \url{https://www.amazon.com/gp/product/B08KRV8VYP}}, significantly cheaper than a force-torque sensor. In practice, it is noteworthy that the cutting friction of different soft materials varies greatly; thus, we manually set the threshold for each material when converting the continuous force signal into a binary collision signal.

\begin{figure}[t]
    \centering
    \includegraphics[width=\linewidth]{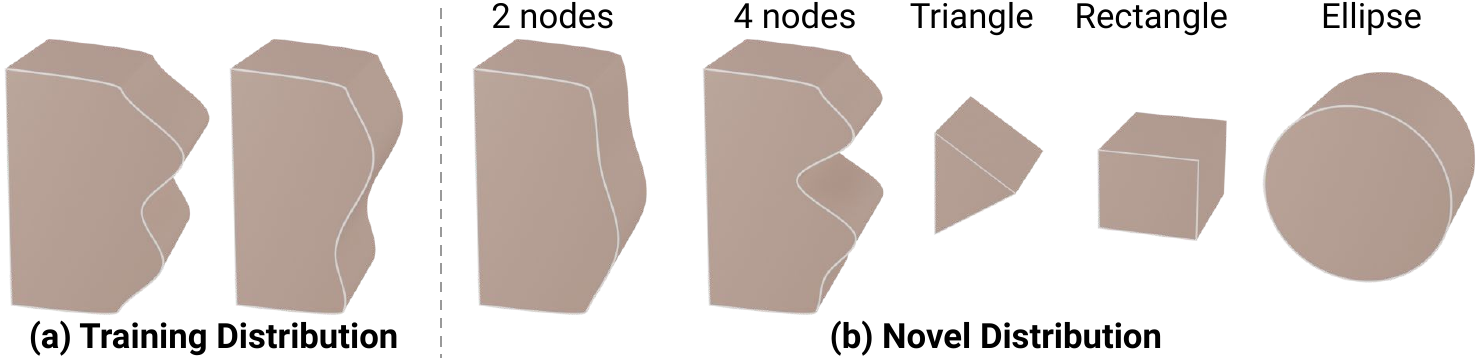}
    
    \caption{\textbf{Generated cores for training and evaluation}}
    \label{fig:sim_core}
    \centering
    \includegraphics[width=\linewidth]{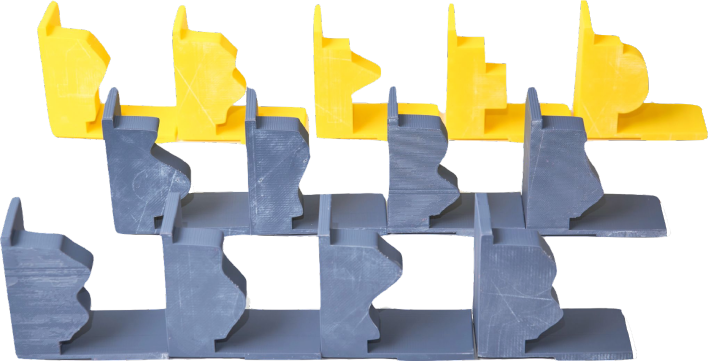}
    \caption{\textbf{3D printed cores in real-world evaluation}. 8 in-distribution geometries (gray) and 5 out-of-distribution geometries  (yellow).}
    \vspace{-4mm}
    \label{fig:real_core}
\end{figure}

\begin{figure*}[t]
    \centering
    \includegraphics[width=0.95\linewidth]{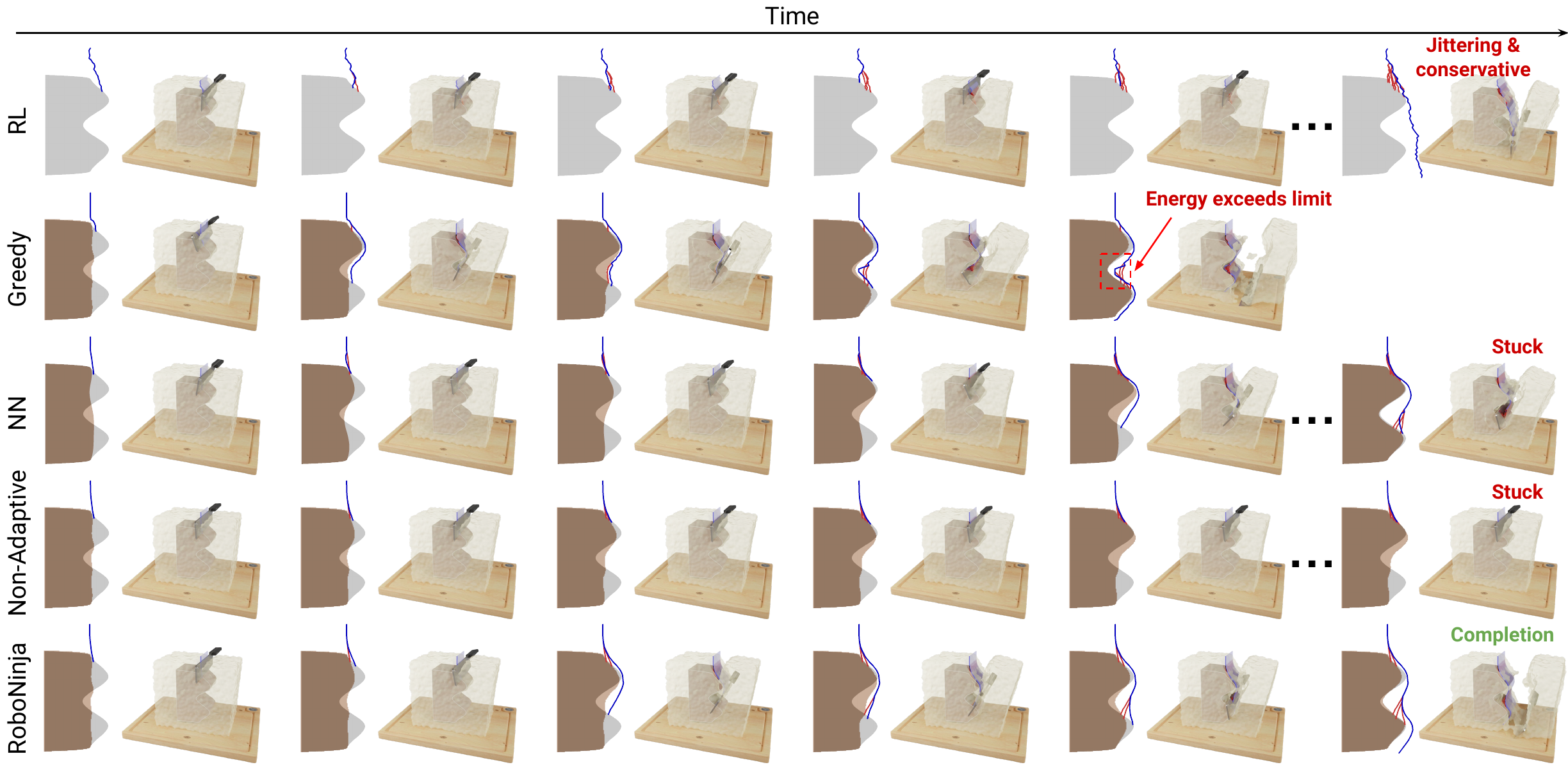}
    \caption{\textbf{Evaluation on in-distribution geometries.} Each column shows an iteration (continuous execution until a collision). In the 2D view  (left), the ground truth geometry is shown in gray, and the estimation is shown in brown. In both views, the forward trajectory of the knife is demonstrated in blue, and the retraction trajectories due to collision are visualized in red. The cutting trajectory of [\RL] is very jittering and becomes too conservative after a few collisions. [\HEURISTIC] strictly follows the contour of the estimated geometry, leading to exceeding energy consumption during abrupt rotations. Both [\NN] and [\OURSNA] can't complete the cutting task within 10 collisions. [\OURS] is able to iteratively update the estimate of the core after each collision and adaptively adjust the cutting trajectory with optimized energy consumption.}
    \vspace{-1mm}
    \label{fig:cut_comparison_sim}
\end{figure*}

 \begin{figure*}[t]
    \centering
    \includegraphics[width=0.96\linewidth]{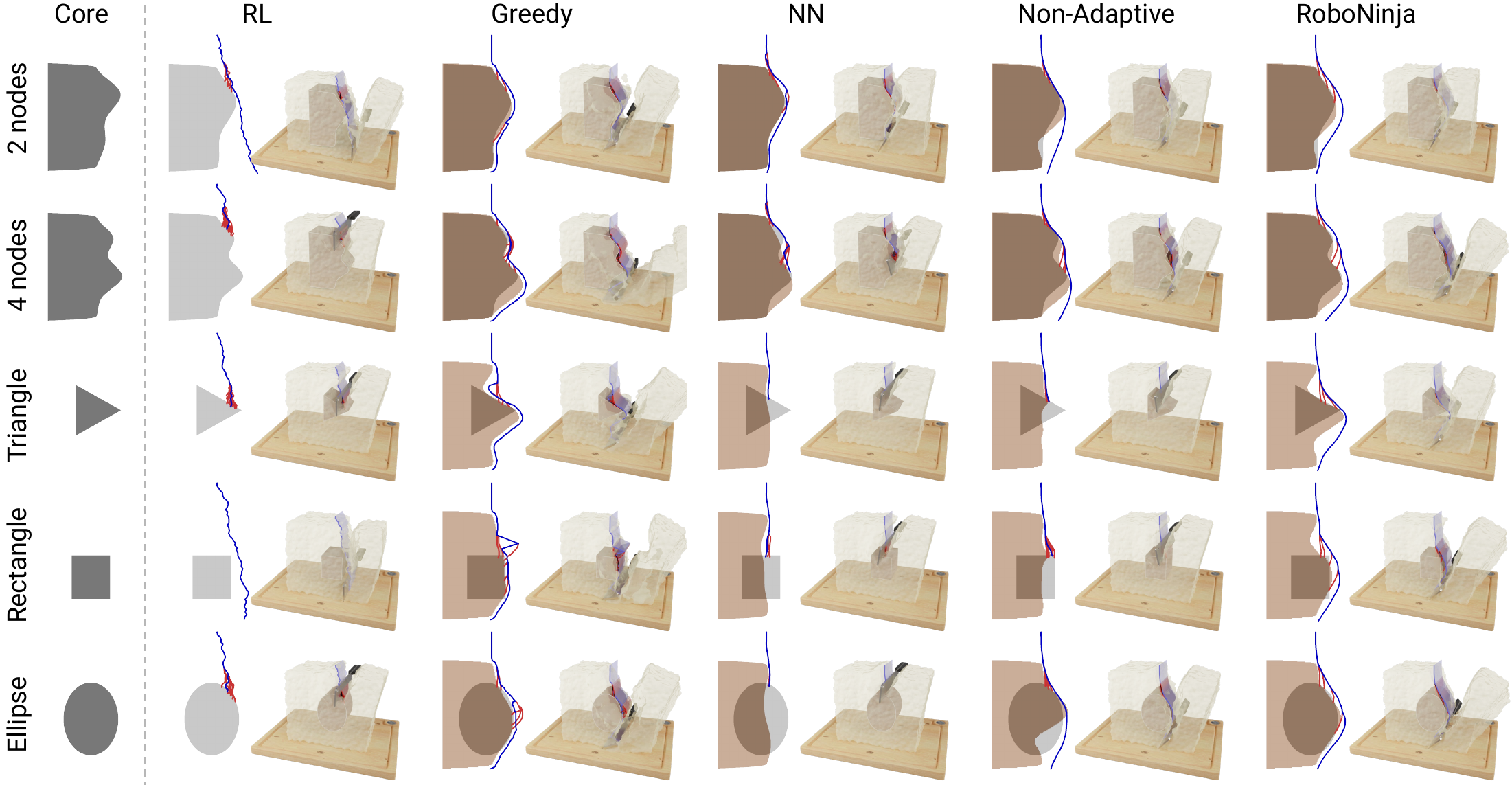}
    \caption{\textbf{Evaluation on out-of-distribution geometries.} The cores sampled from each category are shown in the first column, followed by trajectories produced by each method. Both [\NN] and [\OURSNA] may get stuck, especially when the geometries (e.g., Triangle and Rectangle) are significantly different from the training set. Both [\HEURISTIC] and [\OURS] have a strong generalization ability to handle various novel geometries. However, [\OURS] consumes much less energy than [\HEURISTIC], leading to smoother and more natural cutting trajectories.}
    \vspace{-2mm}
    \label{fig:cut_comparison_sim_ood}
\end{figure*}

\begin{table*}[t]
    \centering
    \setlength\tabcolsep{5pt}
    \begin{tabular}{l|ccccc|ccccc}
    \toprule
    & \multicolumn{5}{c|}{In-distribution Geometries} & \multicolumn{5}{c}{Out-of-distribution Geometries} \\
                & Completion$\uparrow$ & Cut Mass$\uparrow$ & Collision$\downarrow$ & Avg Eng$\downarrow$ & Max Eng$\downarrow$ & Completion$\uparrow$ & Cut Mass$\uparrow$ & Collision$\downarrow$ & Avg Eng$\downarrow$ & Max Eng$\downarrow$ \\
    \midrule
    \RL         & 0.600 & 0.540 & 0.030 & 0.380 & 1.33 & 0.540 & 0.534 & 0.038 & 0.326 & 1.270 \\
    \HEURISTIC  & 0.560 & 0.803 & 0.029 & 0.625 & 2.632 & 0.280 & 0.724 & 0.033 & 0.630 & 2.621 \\
    \NN         & 0.400 & 0.585 & 0.045 & \textbf{0.242} & 1.106 & 0.300 & 0.472 & 0.048 & 0.233 & 1.072 \\
    \OURSNA     & 0.460 & 0.595 & 0.044 & 0.226 & \textbf{0.857} & 0.400 & 0.529 & 0.048 & \textbf{0.224} & \textbf{0.945} \\
    \OURS        & \textbf{1.000} & \textbf{0.875} & \textbf{0.028} & 0.357 & 0.862 & \textbf{0.880} & \textbf{0.799} & 0.033 & 0.342 & 0.988 \\
    
    \bottomrule
    \end{tabular}
    \caption{\textbf{\textbf{Cutting performance on in-distribution and novel geometries.}}}
    \vspace{-4mm}
    \label{tab:simulation_result}
\end{table*}

\section{Evaluation}
\label{sec:evaluation}
In this section, we first evaluate the cutting performance of our proposed method through comparison with various baselines and variants in simulated environments. We further validate our approach by conducting experiments on a real-world setup. Additionally, we conduct ablation studies to evaluate the contribution of each component of our system to its overall performance.

\subsection{Policy Evaluation in Simulation}
\mypara{Dataset and Experimental Setup}
We generate 300 and 100 multi-material objects for training and evaluation, respectively. Each object comprises a rigid core surrounded by soft material. The soft material is simulated as elastoplastic material using the following parameters: $\lambda=1388.89Pa$, $\mu=2083.33 Pa$, $\sigma=200 Pa$, $\rho=10^3kg/m^3$, where $\lambda$ and $\mu$ are Lamé parameters, $\sigma$ the yield stress, and $\rho$ the density. 
The contour of the cores used in training is parameterized by a cubic spline with 3 equally spaced nodes, with 3 degrees of freedom in total. The horizontal coordinate for each node is sampled from a uniform distribution with range [$-0.035m$, $0.035m$].
We subsequently concatenate a fixed back contour and extrude this 2D polygon into a 3D mesh. The 100 cores used in the evaluation are divided into two sets: 50 cores with the same distribution as the training cores and 50 out-of-distribution cores, which consist of 5 categories with 10 cores each: (1) 2 nodes, (2) 4 nodes, (3) triangle, (4) rectangle, (5) ellipse. Examples of the training and testing cores are shown in Fig. \ref{fig:sim_core}.

\mypara{Implementation Details}
For trajectory optimization in the demonstration collection process described in Sec. \ref{method:trajectory_optimization}, we use $\eta_{col} = 2\text{e}4$ and $\eta_e=0.15$. We generate one expert trajectory for each core in the training set, where optimizing each trajectory takes 300 gradient-based updates using the Adam optimizer \cite{adam} with a learning rate of $1\text{e}-2$. Examples of the optimization process are shown in Fig. \ref{fig:optimization_process}.
The state estimation network has an input and output size of 256$\times$256. The classification threshold ($S_{thr}$) is set to 0.3. The training data is generated by randomly sampling 0 to 9 collision points. In the adaptive cutting policy, the number of retraction steps after a collision ($R_{dis}$) is set to 8, and the tolerance increment ($\tau^+$) is set to 0.005. To achieve a smooth trajectory, the tolerance value is linearly decayed to 0 within 5 steps after increasing. Both networks are implemented in PyTorch \cite{paszke2019pytorch} and trained using the Adam optimizer with a learning rate of $1\text{e}-4$ and a weight decay of $1\text{e}-6$. A comprehensive ablation study of some critical parameters is presented in Sec. \ref{sec:ablation}.
\mypara{Metrics.}
We use the following metrics to evaluate the cutting performance:

\begin{itemize}[leftmargin=5mm]
    \item {Completion Rate}. To measure the completion rate of each cutting task, we consider an execution as ``completed'' if the knife reaches the chopping board and as ``failed'' if either the number of collisions exceeds 10 or the energy loss value for any single step exceeds 3.0, as defined in Sec \ref{method:trajectory_optimization}. An episode is terminated as soon as it's considered ``failed'', and all subsequent metrics are evaluated considering only the action sequence executed before the termination.

    \item  {Cut Mass Ratio}. This metric measures the ratio of the removed mass to the total mass of the soft material originally attached to the rigid core. In case of failed execution, the cut mass only considers the soft material on the right side of the cutting trajectory executed till termination. 

    \item  {Collision Ratio}. To account for variations in trajectory length, we normalize the number of collisions by the length of each trajectory.

    \item   {Avg / Max Energy}. We also evaluate the energy consumption averaged over all steps, as well as the maximum energy consumption incurred at a single step during the whole trajectory.
\end{itemize}
\mypara{Baselines.}
We compare our proposed system to the following alternative approaches:
\begin{itemize}[leftmargin=5mm]

\item \RL: A model-free reinforcement learning policy operating within the same observation and action space, as well as using the same collision-retraction mechanism as our method.
This policy is trained using Soft Actor-Critic (SAC) \cite{haarnoja2018soft}, using a dense reward function given by $\mathcal{R} = \mathcal{L}^t_m - \mathcal{L}^{t-1}_m - \eta (\mathcal{L}^{t}_{e} - \mathcal{L}^{t-1}_{e})$, where $\mathcal{L}^t_m$ and $\mathcal{L}^t_e$ are the cumulative cut mass and energy consumption till step $t$, respectively. Upon collision, the cumulative cut mass will decrease due to automatic retraction, resulting in a negative reward, which naturally functions as a collision penalty. The goal of the RL algorithm is to maximize the cumulative reward, which is equivalent to maximizing the total cut mass and meanwhile minimizing the total energy consumption in our setting. Through a grid search over the energy weight $\eta$, we found the best value to be $\eta=0.05$. We train the policy with 3 random seeds and report the best performance.

\item \NN. This approach uses a nearest neighbor cutting policy in place of our adaptive cutting policy. At each step, it retrieves the most similar core from the training set based on the current state estimation result, and executes the closest action step selected from the associated demonstration trajectory based on the current knife pose. 

\item \HEURISTIC. This approach adopts a heuristic policy instead of our adaptive cutting policy. It determines the movement direction and knife rotation at each step in a greedy manner with the aim of maximizing the cut mass while avoiding collision with the estimated core. The primary difference between this method and ours is that it does not factor in energy consumption. 

\item  \OURSNA. This is a non-adaptive variant of our system which uses a fixed tolerance value of 0.
\end{itemize}

\subsection{Results and Analysis}
Qualitative results are summarized in Fig. \ref{fig:cut_comparison_sim} and Fig. \ref{fig:cut_comparison_sim_ood}. Quantitative evaluations are reported in Tab. \ref{tab:simulation_result}.

\mypara{Comparison to model-free RL.} The cutting task is successfully completed by [\RL] in over 50\% of the cases. However, the resulting cut mass is significantly inferior compared to that of [\OURS]. As demonstrated in Fig. \ref{fig:cut_comparison_sim} and \ref{fig:cut_comparison_sim_ood}, the conservative cutting trajectories of [\RL] maintain a significant distance from the bottom part of the core. In contrast, [\OURS] leverages explicit core estimation to follow the contour of the core, resulting in a significantly higher cut mass. Moreover, the cutting trajectories of [\RL] display a noticeable level of jitter, deviating from ideal human-like behavior. Additionally, the knife could get stuck occasionally due to the lack of an explicit adaptive mechanism.

\begin{figure}[t]
    \centering
    \includegraphics[width=\linewidth]{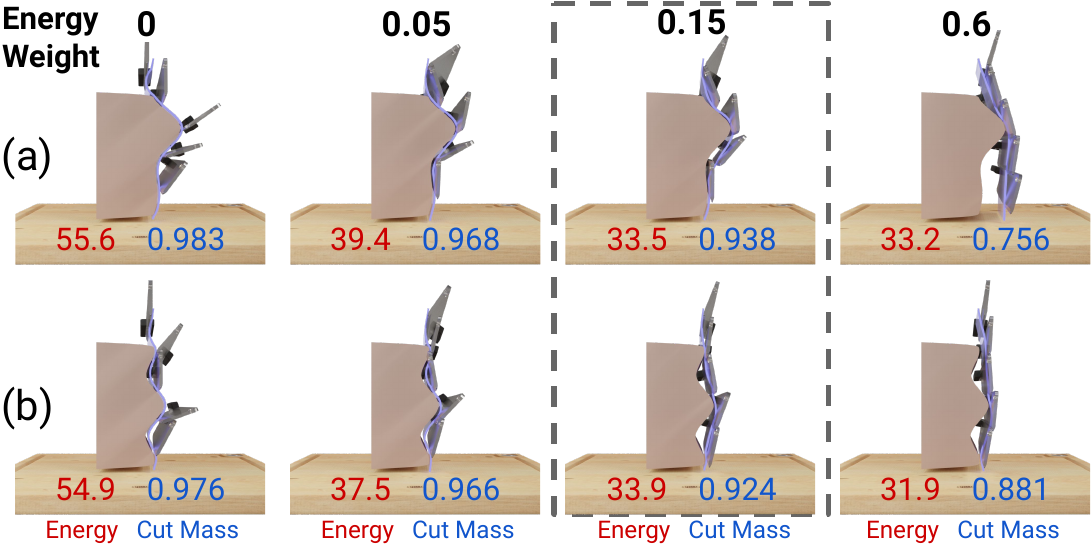}
    
    \caption{\textbf{Trajectories optimized with different weights of energy loss.} Cumulative energy consumption and the cut mass ratio are shown in red and blue, respectively. Without any energy penalty, the trajectory strictly follows the contour of the core and cuts off most of the soft material. However, the knife has to rotate rapidly to avoid collision with the core, which results in large energy consumption. In contrast, a large energy penalty leads to a conservative policy where the rotation of the knife is less noticeable. An appropriate energy weight achieves a balance between energy consumption and cut mass. The one used in our final system [0.15] is able to cut off over 90\% with similar energy consumption as [0.6].}
    \vspace{3mm}
    \label{fig:cut_optimization}
\vspace{1mm}
    \centering
    {\footnotesize
    \setlength\tabcolsep{9pt}
    \begin{tabular}{l|cccc}
    \toprule
    Energy Weight ($\eta_e$)          &0& 0.05& 0.15& 0.6 \\
    \midrule
    Cut Mass Ratio $\uparrow$ & 0.977 & 0.955 & 0.923 & 0.854   \\
    Energy Consumption $\downarrow$ & 53.08 & 36.20 & 32.96 & 30.79   \\
    \bottomrule
    \end{tabular}
    \captionof{table}{\textbf{Effects of different energy weights} \label{tab:optimization_result}}
    \vspace{-5.5mm}
    }
\end{figure}

\mypara{Comparison to \HEURISTIC.}  Compared to [\HEURISTIC], [\OURS] with a learning-based cutting policy achieves around +44\% improvement in terms of completion rate. While [\HEURISTIC] is able to strictly follow the contour of the estimated geometry and maximize the cut mass, the policy does not consider energy consumption. This may result in physically implausible actions, such as abrupt knife rotations. 
In contrast, our cutting policy, which imitates trajectories optimized with the energy consumption objective, is able to sacrifice a small amount of cut mass ratio in exchange for much less energy consumption. This is evident in the energy consumption values in Tab. \ref{tab:optimization_result}, where the average and maximum energy consumption of [\OURS] is 40\% and 60\% less than [\HEURISTIC].

\mypara{Comparison to \NN.} [\NN] directly leverages the action from the demonstration, which reduces the chance of exceeding the energy limit. However, the performance of [\NN] relies heavily on the retrieved nearest neighbor trajectory, making it susceptible to any potential errors in the state estimation result. 
As shown in Fig. \ref{fig:cut_comparison_sim}, in [\NN]'s last step, a slight mismatch between the actual core and the retrieved one results in a collision. Afterward, if the state estimation doesn't have enough change after the collision, the policy will be stuck since the retrieved nearest neighbor remains the same. In contrast, [\OURS] is able to adapt the cutting policy to be more conservative when the state estimation is inaccurate and thereby avoiding being stuck due to collision.

\mypara{Effect of the adaptive cutting policy.} As shown in Fig. \ref{fig:cut_comparison_sim}, [\OURSNA] behaves similarly to [\NN] and suffers from getting stuck at the same location. Thanks to the adaptability of [\OURS], the policy successfully bypasses the peaks of the core with a higher tolerance, which improves the completion rate by over +54\%.

\mypara{Generalization to novel geometries.}
Despite being trained on a single type of core geometry, our policy is able to handle cores with novel geometries thanks to its adaptive cutting policy. Triangles and rectangles are particularly challenging, as they contain straight edges and sharp corners not present in the training data. The qualitative comparison in Fig. \ref{fig:cut_comparison_sim_ood} shows that other baselines may fail due to exceeding energy consumption ([\HEURISTIC]) or getting stuck at one collision location ([\RL], [\NN], and  [\OURSNA]). In contrast, [\OURS] can iteratively update state estimation after the collision and adjust the cutting policy to avoid getting stuck, achieving a balance between energy consumption and the cut mass.

\begin{figure}[t]
    \centering
    \includegraphics[width=\linewidth]{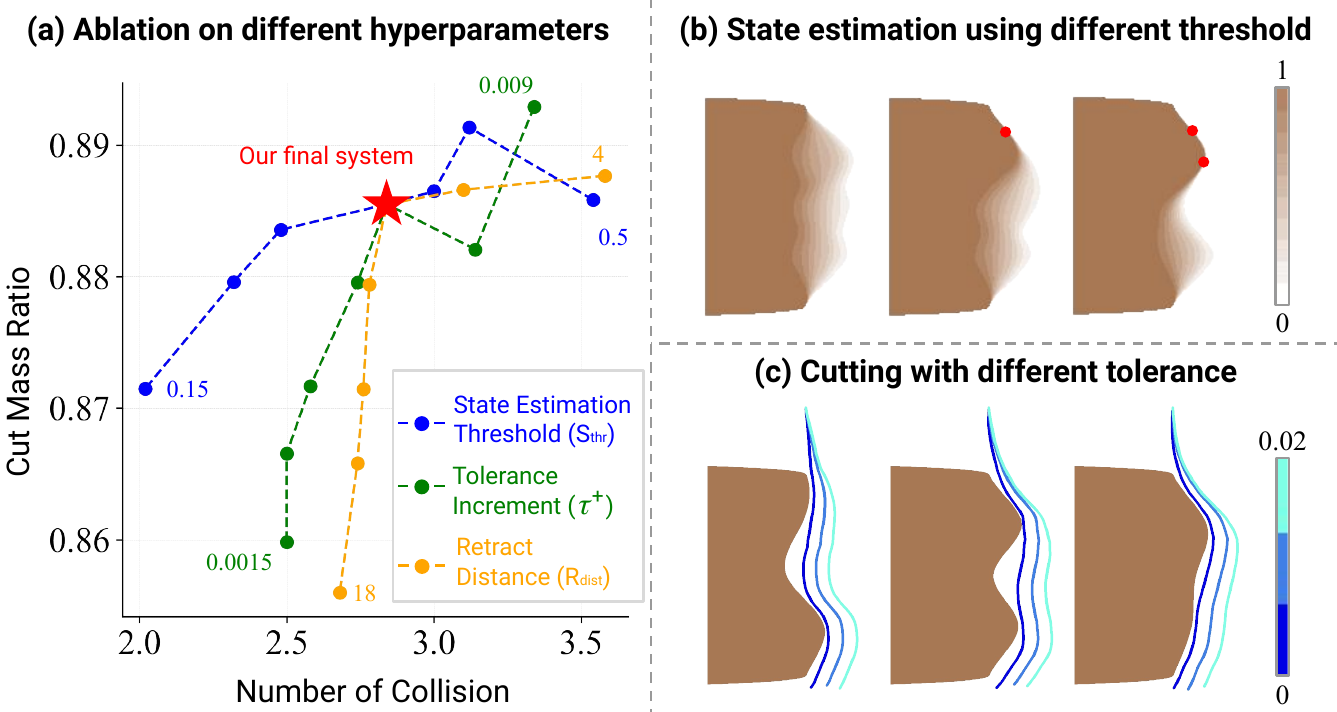}
    
    \caption{\textbf{Ablations on algorithm parameters.} [Left] summarizes the effects of several critical hyperparameters in the algorithm. Detailed discussion about their effects and trade-off can be found in Sec. \ref{sec:ablation}. [Right-Up] State estimation results with different thresholds. [Right-Bottom] Cutting trajectories with different tolerance increments.}
    \vspace{-2mm}
    \label{fig:ablation}
\end{figure}

\begin{figure}[t]
    \centering
    
    {
    \footnotesize
    \setlength\tabcolsep{2.5pt}
    \begin{tabular}{l|ccc|ccc}
    \toprule
    & \multicolumn{3}{c|}{In-distribution Geometries} & \multicolumn{3}{c}{Out-of-distribution Geometries} \\
                & COMP$\uparrow$ & Cut M$\uparrow$ & COLL$\downarrow$ & COMP$\uparrow$ & Cut M$\uparrow$ & COLL$\downarrow$ \\
    \midrule
    
    \OURSNA     & 0.125 & 0.489 & 0.049 & 0.200 & 0.438 & 0.048 \\
    \OURS        & \textbf{1.000} & \textbf{0.877} & \textbf{0.027} & \textbf{1.000} & \textbf{0.824} & \textbf{0.028} \\
    \bottomrule
    \end{tabular}
    \captionof{table}{\textbf{Cutting performance of real-world evaluation.} Here COMP, Cut M, and COLL represent Completion Rate, Cut Mass Ratio, and Collision Ratio respectively.}
    \label{tab:real_result}
    }
    \vspace{2mm}
    \centering
    \includegraphics[width=0.98\linewidth]{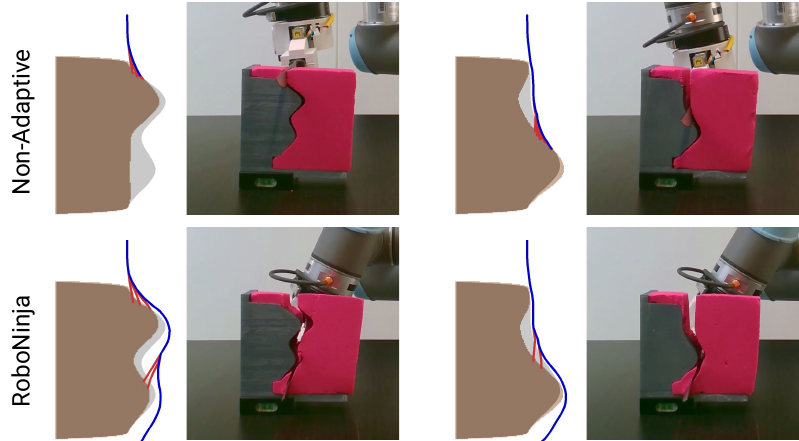}
    \caption{\textbf{Real-world comparison}. [\OURSNA] has the drawback of getting stuck even when the state estimation is very close. In contrast, [\OURS] leverages an adaptive cutting policy that allows it to bypass the peak and successfully reach the bottom.}
    \vspace{-5mm}
    \label{fig:cut_comparison_real}
\end{figure}
\begin{figure*}[t]
    \centering
    \includegraphics[width=0.98\linewidth]{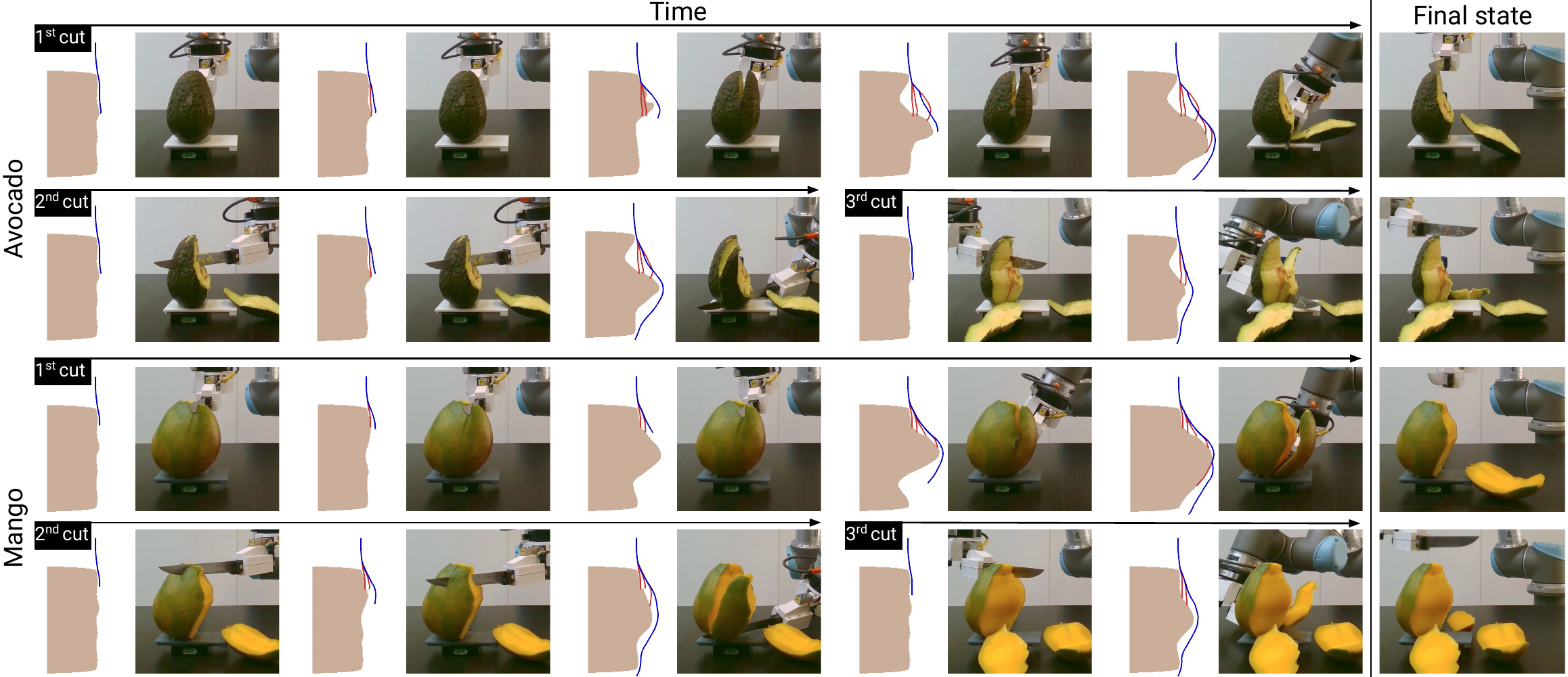}
    \caption{\textbf{Evaluation on real fruits}. The left part illustrates the cutting execution on both an avocado and a mango, with an initial rotation angle of 0\degree, -45\degree, and 45\degree, respectively. The last column displays the final state after one cut and all three cuts.}
    \label{fig:cut_fruit}
    \centering
    \includegraphics[width=0.98\linewidth]{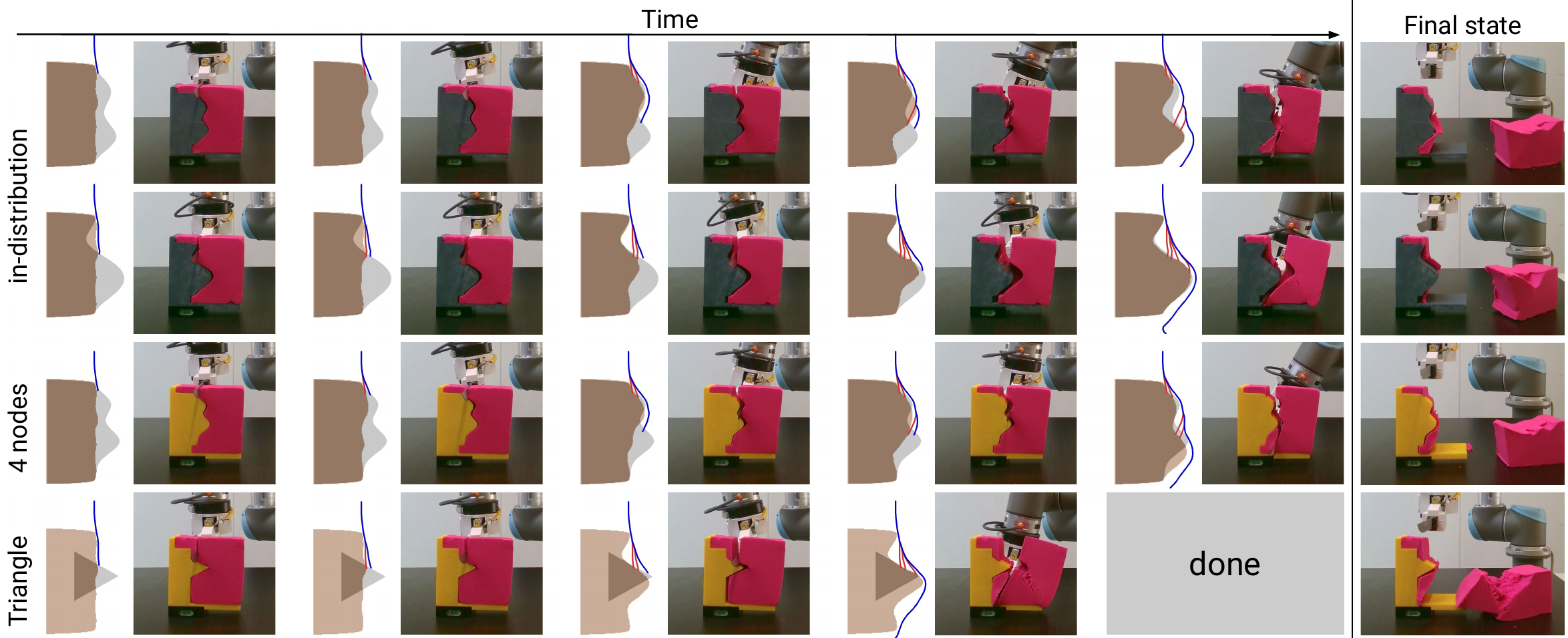}
    \caption{\textbf{Realworld evaluation on 3d printed cores and kinetic sand.} Our simulation-trained policy demonstrates strong generalization capabilities, effectively handling both in-distribution and out-of-distribution cores in a real-world setting. With only a few collisions, it is able to accurately estimate the core geometry and cut off the majority of the sand with a smooth cutting trajectory.}
    \vspace{-5mm}
    \label{fig:cut_real_per_iter}
\end{figure*}
\subsection{Ablation Studies \label{sec:ablation}}
The following experiments study the effects of a few critical parameters and design choices. The experimental results are summarized in Fig. \ref{fig:cut_optimization} and Fig. \ref{fig:ablation}. 

\textbf{Energy weight ($\eta_{e}$)} First, we want to validate the effect of energy penalty on trajectory optimizations. From the qualitative results in Fig. \ref{fig:cut_optimization}, we can observe that the trajectories optimized with no energy penalty strictly follow the contour of the core, but the knife rotation changes frequently to avoid collisions between the knife spine and the core. In contrast, a large energy weight derives an over-conservative cutting behavior. An appropriate energy weight should achieve a good balance between energy consumption and cut mass. The one used in our final system (0.15) is able to cut off over 90\% of the soft material while consuming a comparable amount of energy when compared to the one with a large energy weight.

\textbf{State estimation threshold ($S_{thr}$)} We evaluate our policy with different state estimation thresholds ranging from 0.15 to 0.5. Fig. \ref{fig:ablation} (b) indicates that the estimation results are consistent on the area to the left of the core boundary estimated from collision signals. However, the variation in the threshold value leads to a visible difference in the unexplored area. A small threshold suggests a larger estimated geometry, resulting in a more conservative policy with both fewer collisions and less cut mass (blue line in Fig. \ref{fig:ablation} (a)). We choose 0.3 in our system to achieve a balance between the number of collisions and the cut mass ratio.

\textbf{Tolerance increment ($\tau^+$) } Fig. \ref{fig:ablation} (c) demonstrates the cutting trajectories on the same core with different tolerance values. An aggressive cutting policy with a small tolerance will increase the cut mass but also increase the chance of collisions. As to the tolerance adaptation strategy, the green line in Fig. \ref{fig:ablation} (a) shows that a large tolerance increment value $\tau_+$ results in a conservative policy with fewer collisions and a smaller cut mass.

\textbf{Retract distance ($R_{dis}$ )} Another important design choice is the retract distance, which determines the number of retraction steps after a collision. The yellow line in Fig. \ref{fig:ablation} (a) indicates different trends. From 18 to 8, the cut mass ratio increases significantly with a small increase in the number of collisions, but such benefits no longer exist afterward. Therefore, 8 is selected as the retract distance in the final system.

\subsection{Evaluation on a Real-world Setup}
We directly evaluate the trained model on a real-world platform, where a UR5 robot is equipped with a knife and a force sensor described in Sec. \ref{method:real_setup}.
\mypara{3D printed cores.} In Fig. \ref{fig:real_core}, we 3D print 8 in-distribution and 5 out-of-distribution geometries from the test set. As for the soft material, we use Kinect Sand as a proxy because of its stable physics property. Although a 1-DoF force sensor is sufficient for collision detection, it cannot reflect the energy consumption caused by abrupt rotations. Hence, in our evaluations, we only consider the first three metrics, which are \textit{completion rate}, \textit{cut mass ratio}, and \textit{collision ratio}. For the sake of safety, we exclude [\HEURISTIC] from real-world evaluations and select [\OURSNA] for comparison. The termination criteria for the real-world evaluation are reaching the bottom (completed) or more than 10 collisions (failed). Following the same criteria as in simulation, if the execution is failed, the soft material to the right of the knife's trajectory is considered to be cut off. The quantitative results in Tab. \ref{tab:real_result} and qualitative comparison in Fig. \ref{fig:cut_comparison_real} show that [\OURSNA] gets stuck in most cases. In contrast, [\OURS] is able to bypass the core after a few collisions and complete all test cases. The more detailed qualitative results in Fig. \ref{fig:cut_real_per_iter} demonstrate that [\OURS] is able to accurately estimate core geometry with sparse collision signals only, and cut the soft part off the cores with not only in-distribution but also out-of-distribution geometries in real-world scenarios.

\mypara{Fruits.} We then evaluate our model on real fruits, including avocados, mangos, and plums. To better resemble real-world scenarios, we execute the same policy multiple times with different initial rotation angles about the $y$-axis to cut off more soft material from different directions. Additionally, we augment our vertical cutting trajectory with an additional horizontal and repetitive back-and-forth slicing primitive to effectively cut through fiber-rich material, such as mango\footnote{Note that since the force required to create the initial opening on the mango skin exceeds the force limit of our robot, we had to manually remove a small piece of skin at the top.} and plum skin. Qualitative results on an avocado and a mango are shown in Fig. \ref{fig:cut_fruit}, demonstrating that our model trained with procedurally generated geometries is able to generalize to various fruit cores in the real world. Additionally, the ability to extend the policy allows it to cut off soft material from different directions, making it more practical in the real world.

\begin{figure}[t]
    \centering
    \includegraphics[width=\linewidth]{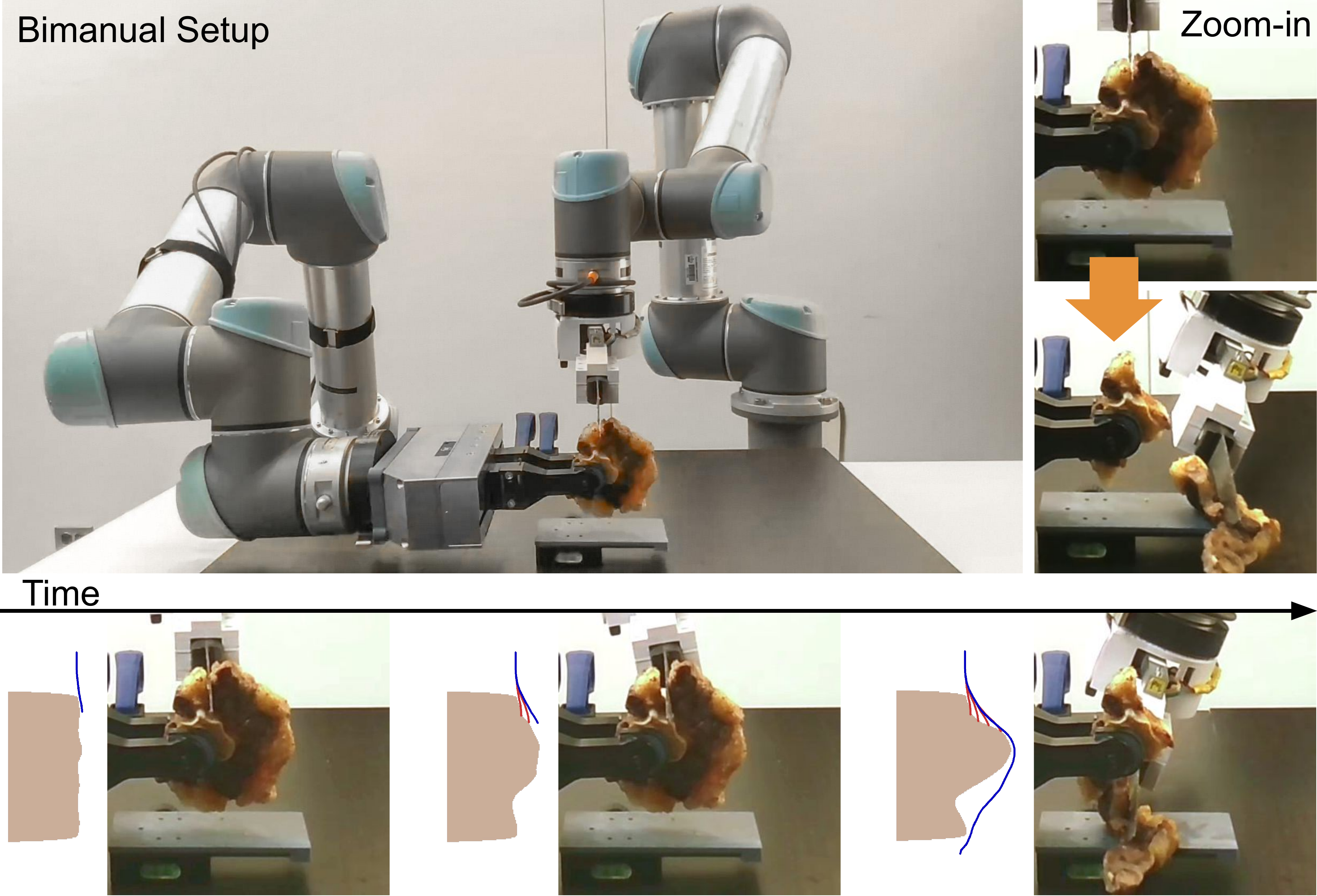}
    \vspace{-5.5mm}
    \caption{\textbf{Evaluation on real meat.} We deploy our cutting policy on a bimanual setup, with one arm using a parallel-jaw gripper to hold the bone and the other arm using a knife to cut the meat. The cutting trajectory and core estimation are shown at the bottom.}
    \vspace{-5.5mm}
    \label{fig:real_meat}
\end{figure}

\mypara{Meat.} We also evaluate our cutting policy on real bone-in meat, specifically, oxtail. In order to resemble real-world situations more realistically, we employ a bimanual setup, where one arm with a parallel-jaw gripper (WSG50) holds the bone and the other arm performs the cutting action. As raw meat is difficult to cut even with the bask-and-forth slicing primitive due to the high collagen content inside tendons, we choose to use cooked meat in this experiment. Qualitative results in Fig. \ref{fig:real_meat} demonstrate \OURS's strong generalization ability in meat-cutting scenarios with novel bone geometry and soft material property.

\subsection{Limitations and Potential Improvements}
There are several limitations and potential improvements of our system:
(a) The knife may exhibit visible deformations (i.e., bending) in real-world scenarios, causing deviations in behavior compared to simulation and disrupting the accuracy of state estimation due to misleading collision positions. This could be addressed by incorporating more accurate knife modeling in our simulator.
(b) In this work, we consider the scenario where the object being cut is secured using an external fixture, and our primary objective is to optimize the trajectory of the cutting tool. In practice, cutting is a bimanual problem, which typically involves coordination between both arms, one to hold and reorient the object and the other to execute the cutting. In the meat-cutting scenario, we attempt to employ such a bimanual setup; however, this setup uses one arm to fix the meat bone and does not fully consider the coordination required between both arms. In addition, human counterparts use a dexterous hand with a soft exterior to firmly hold the object without damaging it, rather than a rigid gripper. Developing such hardware for a dual-arm robotic system, along with the coordinated control policies to perform cutting more efficiently and safely is a promising direction.

\section{Conclusion} 
\label{sec:conclusion}
We introduce \OURS, a robotic system for cutting multi-material objects. The system utilizes demonstrations collected in our newly developed differentiable simulator to train an iterative state estimator and an adaptive cutting policy. It enables the robot to cut soft material off the rigid core while optimizing for both collision occurrences and energy consumption. We also present a low-cost real-world cutting system with real-time force feedback and collision detection, which is used to testify our proposed method on a real robotic setup. Our experiments show that our method is able to generalize well to novel core geometries and even real fruits. We hope our experimental findings and the newly developed simulator could inspire future work on robot learning involving interactions with multi-material objects.

\section*{Acknowledgement} We would like to thank Huy Ha, Zeyi Liu, and Mandi Zhao for their helpful feedback and fruitful discussions. This work was supported in part by  NSF Awards 2037101, 2132519, 2037101, and Toyota Research Institute. Dr. Gan was supported by the DARPA MCS program and gift funding from MERL, Cisco, and Amazon. We would like to thank Google for the UR5 robot hardware. The views and conclusions contained herein are those of the authors and should not be interpreted as necessarily representing the official policies, either expressed or implied, of the sponsors.


\bibliographystyle{plainnat}
\bibliography{references}

\begin{thebibliography}{62}
\providecommand{\natexlab}[1]{#1}
\providecommand{\url}[1]{\texttt{#1}}
\expandafter\ifx\csname urlstyle\endcsname\relax
  \providecommand{\doi}[1]{doi: #1}\else
  \providecommand{\doi}{doi: \begingroup \urlstyle{rm}\Url}\fi

\bibitem[Allen and Michelman(1990)]{allen1990acquisition}
Peter~K Allen and Paul Michelman.
\newblock Acquisition and interpretation of 3-d sensor data from touch.
\newblock 1990.

\bibitem[Areias and Rabczuk(2017)]{areias2017steiner}
P~Areias and Timon Rabczuk.
\newblock Steiner-point free edge cutting of tetrahedral meshes with
  applications in fracture.
\newblock \emph{Finite Elements in Analysis and Design}, 132:\penalty0 27--41,
  2017.

\bibitem[Berndt et~al.(2017)Berndt, Torchelsen, and
  Maciel]{berndt2017efficient}
Iago Berndt, Rafael Torchelsen, and Anderson Maciel.
\newblock Efficient surgical cutting with position-based dynamics.
\newblock \emph{IEEE computer graphics and applications}, 37\penalty0
  (3):\penalty0 24--31, 2017.

\bibitem[Bierbaum et~al.(2008)Bierbaum, Gubarev, and
  Dillmann]{bierbaum2008robust}
Alexander Bierbaum, Ilya Gubarev, and R{\"u}diger Dillmann.
\newblock Robust shape recovery for sparse contact location and normal data
  from haptic exploration.
\newblock In \emph{2008 IEEE/RSJ International Conference on Intelligent Robots
  and Systems}, pages 3200--3205. IEEE, 2008.

\bibitem[Bohg et~al.(2017)Bohg, Hausman, Sankaran, Brock, Kragic, Schaal, and
  Sukhatme]{bohg2017interactive}
Jeannette Bohg, Karol Hausman, Bharath Sankaran, Oliver Brock, Danica Kragic,
  Stefan Schaal, and Gaurav~S Sukhatme.
\newblock Interactive perception: Leveraging action in perception and
  perception in action.
\newblock \emph{IEEE Transactions on Robotics}, 33\penalty0 (6):\penalty0
  1273--1291, 2017.

\bibitem[Childs(2006)]{childs2006friction}
THC Childs.
\newblock Friction modelling in metal cutting.
\newblock \emph{Wear}, 260\penalty0 (3):\penalty0 310--318, 2006.

\bibitem[Chu et~al.(2015)Chu, McMahon, Riano, McDonald, He, Perez-Tejada,
  Arrigo, Darrell, and Kuchenbecker]{chu2015robotic}
Vivian Chu, Ian McMahon, Lorenzo Riano, Craig~G McDonald, Qin He,
  Jorge~Martinez Perez-Tejada, Michael Arrigo, Trevor Darrell, and Katherine~J
  Kuchenbecker.
\newblock Robotic learning of haptic adjectives through physical interaction.
\newblock \emph{Robotics and Autonomous Systems}, 63:\penalty0 279--292, 2015.

\bibitem[Culbertson et~al.(2014)Culbertson, Unwin, and
  Kuchenbecker]{culbertson2014modeling}
Heather Culbertson, Juliette Unwin, and Katherine~J Kuchenbecker.
\newblock Modeling and rendering realistic textures from unconstrained
  tool-surface interactions.
\newblock \emph{IEEE transactions on haptics}, 7\penalty0 (3):\penalty0
  381--393, 2014.

\bibitem[Duenser et~al.(2020)Duenser, Poranne, Thomaszewski, and
  Coros]{duenser2020robocut}
Simon Duenser, Roi Poranne, Bernhard Thomaszewski, and Stelian Coros.
\newblock Robocut: Hot-wire cutting with robot-controlled flexible rods.
\newblock \emph{ACM Transactions on Graphics (TOG)}, 39\penalty0 (4):\penalty0
  98--1, 2020.

\bibitem[Gadre et~al.(2021)Gadre, Ehsani, and Song]{gadre2021act}
Samir~Yitzhak Gadre, Kiana Ehsani, and Shuran Song.
\newblock Act the part: Learning interaction strategies for articulated object
  part discovery.
\newblock \emph{ICCV}, 2021.

\bibitem[Griffith(1921)]{griffith1921vi}
Alan~Arnold Griffith.
\newblock Vi. the phenomena of rupture and flow in solids.
\newblock \emph{Philosophical transactions of the royal society of london.
  Series A, containing papers of a mathematical or physical character},
  221\penalty0 (582-593):\penalty0 163--198, 1921.

\bibitem[Haarnoja et~al.(2018)Haarnoja, Zhou, Abbeel, and
  Levine]{haarnoja2018soft}
Tuomas Haarnoja, Aurick Zhou, Pieter Abbeel, and Sergey Levine.
\newblock Soft actor-critic: Off-policy maximum entropy deep reinforcement
  learning with a stochastic actor.
\newblock In \emph{International conference on machine learning}, pages
  1861--1870. PMLR, 2018.

\bibitem[Heiden et~al.(2021)Heiden, Macklin, Narang, Fox, Garg, and
  Ramos]{heiden2021disect}
Eric Heiden, Miles Macklin, Yashraj~S Narang, Dieter Fox, Animesh Garg, and
  Fabio Ramos.
\newblock {DiSECt: A Differentiable Simulation Engine for Autonomous Robotic
  Cutting}.
\newblock In \emph{Proceedings of Robotics: Science and Systems}, Virtual, July
  2021.
\newblock \doi{10.15607/RSS.2021.XVII.067}.

\bibitem[Hu et~al.(2018)Hu, Fang, Ge, Qu, Zhu, Pradhana, and
  Jiang]{hu2018moving}
Yuanming Hu, Yu~Fang, Ziheng Ge, Ziyin Qu, Yixin Zhu, Andre Pradhana, and
  Chenfanfu Jiang.
\newblock A moving least squares material point method with displacement
  discontinuity and two-way rigid body coupling.
\newblock \emph{ACM Transactions on Graphics (TOG)}, 37\penalty0 (4):\penalty0
  1--14, 2018.

\bibitem[Hu et~al.(2019{\natexlab{a}})Hu, Anderson, Li, Sun, Carr,
  Ragan-Kelley, and Durand]{hu2019difftaichi}
Yuanming Hu, Luke Anderson, Tzu-Mao Li, Qi~Sun, Nathan Carr, Jonathan
  Ragan-Kelley, and Fr{\'e}do Durand.
\newblock Difftaichi: Differentiable programming for physical simulation.
\newblock \emph{arXiv preprint arXiv:1910.00935}, 2019{\natexlab{a}}.

\bibitem[Hu et~al.(2019{\natexlab{b}})Hu, Liu, Spielberg, Tenenbaum, Freeman,
  Wu, Rus, and Matusik]{hu2019chainqueen}
Yuanming Hu, Jiancheng Liu, Andrew Spielberg, Joshua~B Tenenbaum, William~T
  Freeman, Jiajun Wu, Daniela Rus, and Wojciech Matusik.
\newblock Chainqueen: A real-time differentiable physical simulator for soft
  robotics.
\newblock In \emph{2019 International conference on robotics and automation
  (ICRA)}, pages 6265--6271. IEEE, 2019{\natexlab{b}}.

\bibitem[Huang et~al.(2021)Huang, Hu, Du, Zhou, Su, Tenenbaum, and
  Gan]{huang2021plasticinelab}
Zhiao Huang, Yuanming Hu, Tao Du, Siyuan Zhou, Hao Su, Joshua~B Tenenbaum, and
  Chuang Gan.
\newblock Plasticinelab: A soft-body manipulation benchmark with differentiable
  physics.
\newblock \emph{arXiv preprint arXiv:2104.03311}, 2021.

\bibitem[Je{\v{r}}{\'a}bkov{\'a} and Kuhlen(2009)]{jevrabkova2009stable}
Lenka Je{\v{r}}{\'a}bkov{\'a} and Torsten Kuhlen.
\newblock Stable cutting of deformable objects in virtual environments using
  xfem.
\newblock \emph{IEEE computer graphics and applications}, 29\penalty0
  (2):\penalty0 61--71, 2009.

\bibitem[Jiang et~al.(2022)Jiang, Hsu, and Zhu]{jiang2022ditto}
Zhenyu Jiang, Cheng-Chun Hsu, and Yuke Zhu.
\newblock Ditto: Building digital twins of articulated objects from
  interaction.
\newblock In \emph{Proceedings of the IEEE/CVF Conference on Computer Vision
  and Pattern Recognition}, pages 5616--5626, 2022.

\bibitem[Kenney et~al.(2009)Kenney, Buckley, and Brock]{kenney2009interactive}
Jacqueline Kenney, Thomas Buckley, and Oliver Brock.
\newblock Interactive segmentation for manipulation in unstructured
  environments.
\newblock In \emph{2009 IEEE International Conference on Robotics and
  Automation}, pages 1377--1382. IEEE, 2009.

\bibitem[Kingma and Ba(2015)]{adam}
Diederick~P Kingma and Jimmy Ba.
\newblock Adam: A method for stochastic optimization.
\newblock In \emph{International Conference on Learning Representations
  (ICLR)}, 2015.

\bibitem[Koschier et~al.(2014)Koschier, Lipponer, and
  Bender]{koschier2014adaptive}
Dan Koschier, Sebastian Lipponer, and Jan Bender.
\newblock Adaptive tetrahedral meshes for brittle fracture simulation.
\newblock In \emph{Symposium on Computer Animation}, pages 57--66, 2014.

\bibitem[Li et~al.(2018)Li, Wu, Tedrake, Tenenbaum, and
  Torralba]{li2018learning}
Yunzhu Li, Jiajun Wu, Russ Tedrake, Joshua~B Tenenbaum, and Antonio Torralba.
\newblock Learning particle dynamics for manipulating rigid bodies, deformable
  objects, and fluids.
\newblock \emph{arXiv preprint arXiv:1810.01566}, 2018.

\bibitem[Li et~al.(2019)Li, Wu, Zhu, Tenenbaum, Torralba, and
  Tedrake]{li2019propagation}
Yunzhu Li, Jiajun Wu, Jun-Yan Zhu, Joshua~B Tenenbaum, Antonio Torralba, and
  Russ Tedrake.
\newblock Propagation networks for model-based control under partial
  observation.
\newblock In \emph{2019 International Conference on Robotics and Automation
  (ICRA)}, pages 1205--1211. IEEE, 2019.

\bibitem[Lin et~al.(2022{\natexlab{a}})Lin, Huang, Li, Tenenbaum, Held, and
  Gan]{lin2021diffskill}
Xingyu Lin, Zhiao Huang, Yunzhu Li, Joshua~B. Tenenbaum, David Held, and Chuang
  Gan.
\newblock Diffskill: Skill abstraction from differentiable physics for
  deformable object manipulations with tools.
\newblock \emph{International Conference on Learning Representation (ICLR)},
  2022{\natexlab{a}}.

\bibitem[Lin et~al.(2022{\natexlab{b}})Lin, Qi, Zhang, Huang, Fragkiadaki, Li,
  Gan, and Held]{lin2022planning}
Xingyu Lin, Carl Qi, Yunchu Zhang, Zhiao Huang, Katerina Fragkiadaki, Yunzhu
  Li, Chuang Gan, and David Held.
\newblock Planning with spatial-temporal abstraction from point clouds for
  deformable object manipulation.
\newblock In \emph{6th Annual Conference on Robot Learning},
  2022{\natexlab{b}}.

\bibitem[Liu et~al.(2017)Liu, Wu, Sun, and Guo]{liu2017recent}
Huaping Liu, Yupei Wu, Fuchun Sun, and Di~Guo.
\newblock Recent progress on tactile object recognition.
\newblock \emph{International Journal of Advanced Robotic Systems}, 14\penalty0
  (4):\penalty0 1729881417717056, 2017.

\bibitem[Long et~al.(2013)Long, Moughlbay, Khalil, and
  Martinet]{long2013robotic}
Philip Long, Amine Moughlbay, Wisama Khalil, and Philippe Martinet.
\newblock Robotic meat cutting.
\newblock In \emph{Ict-pamm workshop}, 2013.

\bibitem[Lu et~al.(2022)Lu, Wang, and Kumar]{lu2022curiosity}
Yujie Lu, Jianren Wang, and Vikash Kumar.
\newblock Curiosity driven self-supervised tactile exploration of unknown
  objects.
\newblock \emph{arXiv preprint arXiv:2204.00035}, 2022.

\bibitem[Lv et~al.(2022)Lv, Yu, Shao, Liu, Xu, and Lu]{lv2022sagci}
Jun Lv, Qiaojun Yu, Lin Shao, Wenhai Liu, Wenqiang Xu, and Cewu Lu.
\newblock Sagci-system: Towards sample-efficient, generalizable, compositional,
  and incremental robot learning.
\newblock In \emph{2022 International Conference on Robotics and Automation
  (ICRA)}, pages 98--105. IEEE, 2022.

\bibitem[Matsubara and Shibata(2017)]{matsubara2017active}
Takamitsu Matsubara and Kotaro Shibata.
\newblock Active tactile exploration with uncertainty and travel cost for fast
  shape estimation of unknown objects.
\newblock \emph{Robotics and Autonomous Systems}, 91:\penalty0 314--326, 2017.

\bibitem[Merchant(1945)]{merchant1945mechanics}
M~Eugene Merchant.
\newblock Mechanics of the metal cutting process. i. orthogonal cutting and a
  type 2 chip.
\newblock \emph{Journal of applied physics}, 16\penalty0 (5):\penalty0
  267--275, 1945.

\bibitem[Miller et~al.(1999)Miller, Freund, and Needleman]{miller1999modeling}
O~Miller, LB~Freund, and A~Needleman.
\newblock Modeling and simulation of dynamic fragmentation in brittle
  materials.
\newblock \emph{International Journal of Fracture}, 96\penalty0 (2):\penalty0
  101--125, 1999.

\bibitem[Mu et~al.(2019)Mu, Xue, and Jia]{mu2019robotic}
Xiaoqian Mu, Yuechuan Xue, and Yan-Bin Jia.
\newblock Robotic cutting: Mechanics and control of knife motion.
\newblock In \emph{2019 International Conference on Robotics and Automation
  (ICRA)}, pages 3066--3072. IEEE, 2019.

\bibitem[Nie et~al.(2022)Nie, Gadre, Ehsani, and Song]{nie2022sfa}
Neil Nie, Samir~Yitzhak Gadre, Kiana Ehsani, and Shuran Song.
\newblock Structure from action: Learning interactions for articulated object
  3d structure discovery.
\newblock \emph{arxiv}, 2022.

\bibitem[Pan et~al.(2015)Pan, Bai, Zhao, Hao, and Qin]{pan2015real}
Junjun Pan, Junxuan Bai, Xin Zhao, Aimin Hao, and Hong Qin.
\newblock Real-time haptic manipulation and cutting of hybrid soft tissue
  models by extended position-based dynamics.
\newblock \emph{Computer Animation and Virtual Worlds}, 26\penalty0
  (3-4):\penalty0 321--335, 2015.

\bibitem[Paszke et~al.(2019)Paszke, Gross, Massa, Lerer, Bradbury, Chanan,
  Killeen, Lin, Gimelshein, Antiga, et~al.]{paszke2019pytorch}
Adam Paszke, Sam Gross, Francisco Massa, Adam Lerer, James Bradbury, Gregory
  Chanan, Trevor Killeen, Zeming Lin, Natalia Gimelshein, Luca Antiga, et~al.
\newblock Pytorch: An imperative style, high-performance deep learning library.
\newblock \emph{Advances in neural information processing systems}, 32, 2019.

\bibitem[Paulus et~al.(2015)Paulus, Untereiner, Courtecuisse, Cotin, and
  Cazier]{paulus2015virtual}
Christoph~J Paulus, Lionel Untereiner, Hadrien Courtecuisse, St{\'e}phane
  Cotin, and David Cazier.
\newblock Virtual cutting of deformable objects based on efficient topological
  operations.
\newblock \emph{The Visual Computer}, 31\penalty0 (6):\penalty0 831--841, 2015.

\bibitem[Pfaff et~al.(2020)Pfaff, Fortunato, Sanchez-Gonzalez, and
  Battaglia]{pfaff2020learning}
Tobias Pfaff, Meire Fortunato, Alvaro Sanchez-Gonzalez, and Peter~W Battaglia.
\newblock Learning mesh-based simulation with graph networks.
\newblock \emph{arXiv preprint arXiv:2010.03409}, 2020.

\bibitem[Qi et~al.(2022)Qi, Lin, and Held]{qi2022dough}
Carl Qi, Xingyu Lin, and David Held.
\newblock Learning closed-loop dough manipulation using a differentiable reset
  module.
\newblock \emph{IEEE Robotics and Automation Letters}, pages 1--8, 2022.
\newblock \doi{10.1109/LRA.2022.3191239}.

\bibitem[Sawhney et~al.(2021)Sawhney, Lee, Zhang, Veloso, and
  Kroemer]{sawhney2021playing}
Amrita Sawhney, Steven Lee, Kevin Zhang, Manuela Veloso, and Oliver Kroemer.
\newblock Playing with food: Learning food item representations through
  interactive exploration.
\newblock In \emph{International Symposium on Experimental Robotics}, pages
  309--322. Springer, 2021.

\bibitem[Schiebener et~al.(2011)Schiebener, Ude, Morimoto, Asfour, and
  Dillmann]{schiebener2011segmentation}
David Schiebener, Ale{\v{s}} Ude, Jun Morimoto, Tamim Asfour, and R{\"u}diger
  Dillmann.
\newblock Segmentation and learning of unknown objects through physical
  interaction.
\newblock In \emph{2011 11th IEEE-RAS International Conference on Humanoid
  Robots}, pages 500--506. IEEE, 2011.

\bibitem[Schiebener et~al.(2013)Schiebener, Morimoto, Asfour, and
  Ude]{schiebener2013integrating}
David Schiebener, Jun Morimoto, Tamim Asfour, and Ale{\v{s}} Ude.
\newblock Integrating visual perception and manipulation for autonomous
  learning of object representations.
\newblock \emph{Adaptive Behavior}, 21\penalty0 (5):\penalty0 328--345, 2013.

\bibitem[Schmitz et~al.(2014)Schmitz, Bansho, Noda, Iwata, Ogata, and
  Sugano]{schmitz2014tactile}
Alexander Schmitz, Yusuke Bansho, Kuniaki Noda, Hiroyasu Iwata, Tetsuya Ogata,
  and Shigeki Sugano.
\newblock Tactile object recognition using deep learning and dropout.
\newblock In \emph{2014 IEEE-RAS International Conference on Humanoid Robots},
  pages 1044--1050. IEEE, 2014.

\bibitem[Schneider et~al.(2009)Schneider, Sturm, Stachniss, Reisert, Burkhardt,
  and Burgard]{schneider2009object}
Alexander Schneider, J{\"u}rgen Sturm, Cyrill Stachniss, Marco Reisert, Hans
  Burkhardt, and Wolfram Burgard.
\newblock Object identification with tactile sensors using bag-of-features.
\newblock In \emph{2009 IEEE/RSJ International Conference on Intelligent Robots
  and Systems}, pages 243--248. IEEE, 2009.

\bibitem[Sharma et~al.(2019)Sharma, Zhang, and Kroemer]{sharma2019learning}
Mohit Sharma, Kevin Zhang, and Oliver Kroemer.
\newblock Learning semantic embedding spaces for slicing vegetables.
\newblock \emph{arXiv preprint arXiv:1904.00303}, 2019.

\bibitem[S{\o}ndergaard et~al.(2016)S{\o}ndergaard, Feringa, N{\o}rbjerg,
  Steenstrup, Brander, Graversen, Markvorsen, B{\ae}rentzen, Petkov, Hattel,
  et~al.]{sondergaard2016robotic}
Asbj{\o}rn S{\o}ndergaard, Jelle Feringa, Toke N{\o}rbjerg, Kasper Steenstrup,
  David Brander, Jens Graversen, Steen Markvorsen, Andreas B{\ae}rentzen, Kiril
  Petkov, Jesper Hattel, et~al.
\newblock Robotic hot-blade cutting.
\newblock In \emph{Robotic fabrication in architecture, art and design 2016},
  pages 150--164. Springer, 2016.

\bibitem[Stomakhin et~al.(2013)Stomakhin, Schroeder, Chai, Teran, and
  Selle]{stomakhin2013material}
Alexey Stomakhin, Craig Schroeder, Lawrence Chai, Joseph Teran, and Andrew
  Selle.
\newblock A material point method for snow simulation.
\newblock \emph{ACM Transactions on Graphics (TOG)}, 32\penalty0 (4):\penalty0
  1--10, 2013.

\bibitem[Wang et~al.(2019)Wang, Ding, Gast, Zhu, Gagniere, Jiang, and
  Teran]{wang2019simulation}
Stephanie Wang, Mengyuan Ding, Theodore~F Gast, Leyi Zhu, Steven Gagniere,
  Chenfanfu Jiang, and Joseph~M Teran.
\newblock Simulation and visualization of ductile fracture with the material
  point method.
\newblock \emph{Proceedings of the ACM on Computer Graphics and Interactive
  Techniques}, 2\penalty0 (2):\penalty0 1--20, 2019.

\bibitem[Wang et~al.(2023)Wang, Spielberg, Ma, Xian, Zhang, Tenenbaum, and
  Gan]{wang2023softzoo}
Tsun-Hsuan Wang, Andrew~Everett Spielberg, Pingchuan Ma, Zhou Xian, Hao Zhang,
  Joshua~B. Tenenbaum, and Chuang Gan.
\newblock Softzoo: A soft robot co-design benchmark for locomotion in diverse
  environments.
\newblock In \emph{International Conference on Learning Representations}, 2023.

\bibitem[Wolper et~al.(2019)Wolper, Fang, Li, Lu, Gao, and Jiang]{wolper2019cd}
Joshuah Wolper, Yu~Fang, Minchen Li, Jiecong Lu, Ming Gao, and Chenfanfu Jiang.
\newblock Cd-mpm: continuum damage material point methods for dynamic fracture
  animation.
\newblock \emph{ACM Transactions on Graphics (TOG)}, 38\penalty0 (4):\penalty0
  1--15, 2019.

\bibitem[Wolper et~al.(2020)Wolper, Chen, Li, Fang, Qu, Lu, Cheng, and
  Jiang]{wolper2020anisompm}
Joshuah Wolper, Yunuo Chen, Minchen Li, Yu~Fang, Ziyin Qu, Jiecong Lu, Meggie
  Cheng, and Chenfanfu Jiang.
\newblock Anisompm: Animating anisotropic damage mechanics.
\newblock \emph{ACM Trans. Graph.}, 39\penalty0 (4), 2020.

\bibitem[Wu et~al.(2015)Wu, Westermann, and Dick]{wu2015survey}
Jun Wu, R{\"u}diger Westermann, and Christian Dick.
\newblock A survey of physically based simulation of cuts in deformable bodies.
\newblock In \emph{Computer Graphics Forum}, volume~34, pages 161--187. Wiley
  Online Library, 2015.

\bibitem[Xian et~al.(2023)Xian, Zhu, Xu, Tung, Torralba, Fragkiadaki, and
  Gan]{xian2023fluidlab}
Zhou Xian, Bo~Zhu, Zhenjia Xu, Hsiao-Yu Tung, Antonio Torralba, Katerina
  Fragkiadaki, and Chuang Gan.
\newblock Fluidlab: A differentiable environment for benchmarking complex fluid
  manipulation.
\newblock In \emph{International Conference on Learning Representations}, 2023.

\bibitem[Xu et~al.(2022{\natexlab{a}})Xu, Makoviychuk, Narang, Ramos, Matusik,
  Garg, and Macklin]{xu2022accelerated}
Jie Xu, Viktor Makoviychuk, Yashraj Narang, Fabio Ramos, Wojciech Matusik,
  Animesh Garg, and Miles Macklin.
\newblock Accelerated policy learning with parallel differentiable simulation.
\newblock \emph{arXiv preprint arXiv:2204.07137}, 2022{\natexlab{a}}.

\bibitem[Xu et~al.(2022{\natexlab{b}})Xu, Lin, Song, and
  Ciocarlie]{xu2022tandem3d}
Jingxi Xu, Han Lin, Shuran Song, and Matei Ciocarlie.
\newblock Tandem3d: Active tactile exploration for 3d object recognition.
\newblock \emph{arXiv preprint arXiv:2209.08772}, 2022{\natexlab{b}}.

\bibitem[Xu et~al.(2022{\natexlab{c}})Xu, Song, and Ciocarlie]{xu2022tandem}
Jingxi Xu, Shuran Song, and Matei Ciocarlie.
\newblock Tandem: Learning joint exploration and decision making with tactile
  sensors.
\newblock \emph{IEEE Robotics and Automation Letters}, 2022{\natexlab{c}}.

\bibitem[Xu et~al.(2020)Xu, He, Wu, and Song]{xu2020learning}
Zhenjia Xu, Zhanpeng He, Jiajun Wu, and Shuran Song.
\newblock Learning 3d dynamic scene representations for robot manipulation.
\newblock In \emph{Conference on Robot Learning (CoRL)}, 2020.

\bibitem[Zhang et~al.(2019)Zhang, Sharma, Veloso, and
  Kroemer]{zhang2019leveraging}
Kevin Zhang, Mohit Sharma, Manuela Veloso, and Oliver Kroemer.
\newblock Leveraging multimodal haptic sensory data for robust cutting.
\newblock In \emph{2019 IEEE-RAS 19th International Conference on Humanoid
  Robots (Humanoids)}, pages 409--416. IEEE, 2019.

\bibitem[Zhou et~al.(2006{\natexlab{a}})Zhou, Claffee, Lee, and
  McMurray]{zhou2006cutting1}
Debao Zhou, Mark~R Claffee, Kok-Meng Lee, and Gary~V McMurray.
\newblock Cutting," by pressing and slicing", applied to robotic cutting
  bio-materials. i. modeling of stress distribution.
\newblock In \emph{Proceedings 2006 IEEE International Conference on Robotics
  and Automation, 2006. ICRA 2006.}, pages 2896--2901. IEEE,
  2006{\natexlab{a}}.

\bibitem[Zhou et~al.(2006{\natexlab{b}})Zhou, Claffee, Lee, and
  McMurray]{zhou2006cutting2}
Debao Zhou, Mark~R Claffee, Kok-Meng Lee, and Gary~V McMurray.
\newblock Cutting,'by pressing and slicing', applied to the robotic cut of
  bio-materials. ii. force during slicing and pressing cuts.
\newblock In \emph{Proceedings 2006 IEEE International Conference on Robotics
  and Automation, 2006. ICRA 2006.}, pages 2256--2261. IEEE,
  2006{\natexlab{b}}.

\bibitem[Zhou et~al.(2005)Zhou, Molinari, and Shioya]{zhou2005rate}
Fenghua Zhou, Jean-Francois Molinari, and Tadashi Shioya.
\newblock A rate-dependent cohesive model for simulating dynamic crack
  propagation in brittle materials.
\newblock \emph{Engineering fracture mechanics}, 72\penalty0 (9):\penalty0
  1383--1410, 2005.

\end{thebibliography}

\end{document}